\documentclass{article}

\usepackage{microtype}
\usepackage{graphicx}
\usepackage{caption}
\usepackage{subcaption}
\usepackage{booktabs} % for professional tables
\usepackage{amsfonts}
\usepackage{hyperref}
\usepackage{amsmath}
\usepackage{cleveref}
\usepackage{multirow}
\usepackage{amsthm}
\usepackage{xcolor}

\usepackage[accepted]{icml2021}

\icmltitlerunning{On Out-of-distribution Detection with Energy-Based Models}

\begin{document}
\frenchspacing

\twocolumn[
\icmltitle{On Out-of-distribution Detection with Energy-based Models}



\icmlsetsymbol{equal}{*}

\begin{icmlauthorlist}
\icmlauthor{Sven Elflein}{tum}
\icmlauthor{Bertrand Charpentier}{tum}
\icmlauthor{Daniel Z\"ugner}{tum}
\icmlauthor{Stephan G\"unnemann}{tum}
\end{icmlauthorlist}

\icmlaffiliation{tum}{Technical University Munich, Germany}

\icmlcorrespondingauthor{Sven Elflein}{sven.elflein@in.tum.de}

\icmlkeywords{Machine Learning, ICML}

\vskip 0.3in
]



\printAffiliationsAndNotice{}  % leave blank if no need to mention equal contribution

\begin{abstract}
Several density estimation methods have shown to fail to detect out-of-distribution (OOD) samples by assigning higher likelihoods to anomalous data. Energy-based models (EBMs) are flexible, unnormalized density models which seem to be able to improve upon this failure mode. In this work, we provide an extensive study investigating OOD detection with EBMs trained with different approaches on tabular and image data and find that EBMs do not provide consistent advantages. We hypothesize that EBMs do not learn semantic features despite their discriminative structure similar to Normalizing Flows. To verify this hypotheses, we show that supervision and architectural restrictions improve the OOD detection of EBMs independent of the training approach.
\end{abstract}

\section{Introduction}
\label{ch:introduction}

% \begin{itemize}
%     \item Machine learning in more fields, also security critical applications
%     \item Open question how to guarantuee that deep learning models account for samples away from the training distribution
%     \item One option to learn the underlying density of training data
%     \item Normalizing flows assign higher likelihood to OOD data
%     \item EBMs provide more flexible density estimator and has shown that can outperform normalizing flows on image generation
%     \item However no exact maximum likelihood training of EBMs possible  
%     \item Resulting in different training approaches which do only provide either limited or no results and comparison w.r.t. OOD detection
%     \item This work aims to fill this gap and jointly investigates different training approaches under the same setup.
%     \item We further consider the influence of supervision, dimensionality reduction and architectural changes encouraging higher level features.
% \end{itemize}

To leverage deep learning in security-critical application areas, such as medical applications and autonomous driving, robustness, and uncertainty have recently received increased attention \cite{varshneyEngineeringSafetyMachine2016}. 
An open question is to ensure that the model does account for out-of-distribution (OOD) data where recent work has shown that models tend to make over-confident predictions \cite{lakshminarayananSimpleScalablePredictive2017, heinWhyReLUNetworks2019}.
One approach to solve this problem is to estimate the training data distribution and reject samples if the density at that point is low. However, Normalizing Flows, \cite{rezendeVariationalInferenceNormalizing2016} which are powerful density estimators based on a sequence of invertible transformations, tend to assign higher likelihoods to the OOD than the in-distribution (ID) data \cite{nalisnickDeepGenerativeModels2019}.
Another promising class of density estimators without restrictions on the architecture are Energy-based models (EBM) \cite{lecunTutorialEnergyBasedLearning}. Recently, \citet{grathwohlYourClassifierSecretly2020} improved OOD detection by interpreting discriminative models as EBMs.
This encourages that EBMs might be better suited for the task of OOD detection. In this work, we aim to investigate this claim and the main factors facilitating superior OOD detection of EBMs. \\
We summarize our contributions as follows: \textbf{(1)} we find that EBMs do not strictly outperform Normalizing Flows across multiple training methods, %\bc{At this point of the reading, it is not clear to me what 'structured inputs' means.} 
\textbf{(2)} identify that learning semantic features induced by supervision improves OOD detection in recent discriminative EBMs \cite{grathwohlYourClassifierSecretly2020}
and, \textbf{(3)} show that one can use architectural modifications to improve OOD detection with EBMs similar to Normalizing Flows \cite{kirichenkoWhyNormalizingFlows2020}. 
%\bc{it might be relevant to say what NF does it these cases.}

\section{Related Work}
\label{ch:related_work}

\textbf{Classifier-based OOD detection.}
Initially, \citet{hendrycksBaselineDetectingMisclassified2018} proposed to use the maximum softmax probability as OOD score. \citet{liangEnhancingReliabilityOutofdistribution2020, hsuGeneralizedODINDetecting2020} augment this approach by temperature scaling. Other methods add additional loss terms to the objective to encourage maximum entropy predictions for OOD inputs \cite{hendrycksDeepAnomalyDetection2019, leeTrainingConfidencecalibratedClassifiers2018, sricharanBuildingRobustClassifiers2018, heinWhyReLUNetworks2019}.
\citet{malininPredictiveUncertaintyEstimation2018, malininReverseKLDivergenceTraining2019, charpentierPosteriorNetworkUncertainty2020} obtain uncertainty estimates for OOD detection by predicting parameters of a Dirichlet distribution for classification. \\ % Further methods obtain uncertainty estimates based on a Bayesian or approximately Bayesian framework \cite{malininPredictiveUncertaintyEstimation2018, lakshminarayananSimpleScalablePredictive2017}.
\textbf{Density-based OOD detection.}
A set of methods estimates the distribution over activations at multiple layers \cite{leeSimpleUnifiedFramework2018a, zisselmanDeepResidualFlow2020}.
Other methods focus on the data distribution directly: \citet{nalisnickDeepGenerativeModels2019} discovered that the density learned by generative models cannot distinguish between ID and OOD inputs. Various works study this observation identifying background statistic \cite{renLikelihoodRatiosOutofDistribution2019}, excessive influence of input complexity \cite{serraInputComplexityOutofdistribution2020}, and mismatch between the typical set and high-density regions \cite{nalisnickDetectingOutofDistributionInputs2019, choiWAICWhyGenerative2019, morningstarDensityStatesEstimation2020} as causes. In comparison to our work, these methods focus on flow-based and autoregressive density methods with tractable likelihood. \\
Recently, there has also been increasing interest in leveraging EBMs as generative models for OOD detection. 
\citet{duImplicitGenerationGeneralization2020b} investigate the generative capabilities and generalization of EBMs to OOD inputs.
\citet{zhaiDeepStructuredEnergy2016} train EBM architectures with a score matching objective for anomaly detection.
\citet{grathwohlYourClassifierSecretly2020, grathwohlNoMCMCMe2020} derive optimization procedures for hybrid EBMs and investigate their OOD detection performance.
However, existing work does not study the factors leading to improved OOD detection with EBMs compared to other generative models.
Thus, most relevant to our work are the studies by \citet{kirichenkoWhyNormalizingFlows2020, schirrmeisterUnderstandingAnomalyDetection2020} which found that Normalizing Flows learn low-level features common to image datasets and thus struggle with detecting OOD inputs. We aim to provide similar insight for EBMs. 
% \bc{I feel that the related work could be richer. There would be some interesting references in the 'Task-Specific OOD' paragraph of \url{https://arxiv.org/pdf/2105.04471.pdf} }

\section{Method}
\label{ch:method}
In the following, we specify the structure of the EBMs and provide an overview of the training methods considered in this work. \\
\textbf{Energy-based model.}
EBMs \cite{lecunTutorialEnergyBasedLearning} are defined by an energy-function \( E_\theta \) which defines a density over the data \(x\) as

\begin{equation}
    \label{eq:ebm}
    \smash{p_\theta(x) = \frac{\exp(-E_\theta(x))}{Z(\theta)}}
\end{equation}

where \(\smash{Z(\theta) = \int \exp(-E_\theta(x)) dx}\) is the normalizing constant and \(\theta\) are learnable parameters. In particular, \( E_\theta \) can be any function \( E: \mathbb{R}^D \mapsto \mathbb{R} \) placing no restrictions on the model compared to Normalizing Flows. \\
\textbf{Joint Energy model.}
We additionally consider Joint Energy models (JEM) for discriminative EBMs \cite{grathwohlYourClassifierSecretly2020}. Given a classifier \( f : \mathbb{R}^D \mapsto \mathbb{R}^C \) assigning logits for \(C\) classes for a datapoint \(x \in \mathbb{R}^D\), the probabilities over the classes are defined as

\begin{equation}
    \label{eq:jem_p_y_given_x}
    \smash{p_\theta(y \mid x) = \frac{\exp(f_\theta(x)[y])}{\sum_{y^\prime} \exp(f_\theta(x)[y^\prime]}}
\end{equation}

where \( f_\theta(x)[y] \) denotes the \(y\)-th logit.
The logits \( f_\theta(x)[y] \) can be interpret as unnormalized probabilities of the joint distribution \(p_\theta(x, y)\)
%
% \begin{equation}
%     p_\theta(x, y) = \frac{\exp(f(x)[y])}{Z_\theta}
% \end{equation}
which yields the marginal distribution over \(x\) as

\begin{equation}
    \label{eq:jem_p_x}
    p_\theta(x) = \sum_y p_\theta (x, y) = \sum_y \frac{\exp(f(x)[y])}{Z(\theta)}
\end{equation}

% \subsection{Training}

We follow \cite{grathwohlYourClassifierSecretly2020} and optimize the factorization 
\(\smash{\log p_\theta(x, y) =  \log p_\theta(x) + \log p_\theta(y \mid x)}\)
using \Cref{eq:jem_p_y_given_x} and \Cref{eq:jem_p_x}. In particular, we use a Cross Entropy objective to optimize \(\smash{p_\theta(y \mid x)}\) weighted with hyperparameter $\gamma$. \\
For optimizing \(p_\theta(x)\), we consider different approaches which have shown to scale to high-dimensional data. Note that this term should contribute significantly to the OOD detection performance of the model. We introduce the training approaches used in this work in the following. \\
%\bc{I feel that we could underline more that the p(x) optimization is the key part here} 
%\bc{I like when there are small efficient paragraph but it might save some space to merge them a bit more.}
%
\textbf{Sliced score matching.}
\citet{hyvarinenEstimationNonNormalizedStatistical2005} propose to learn an unnormalized density by approximating the score of the distribution \(s_\theta(x) = \nabla_x p(x)\).
% \bc{Is there a problem with the next equation ? The difference between log and non log looks weird}
% \begin{equation}
%     L(\theta) = \frac{1}{2} \mathbb{E}_{p(x)} \left[ \| \nabla_x \log p_\theta(x) - \nabla_x \log p(x)  \| \right]
% \end{equation}
%
% and proposes an optimization scheme requiring only samples from the groundtruth distribution using integration by parts. While this objective does not depend on the normalizing constant \( Z(\theta) \), the optimization involves matrix Hessian-products which are expensive to compute for high dimensional data.
\citet{songSlicedScoreMatching2019} introduce an efficient update formula based on random projection 
\(
    \smash{
    \mathbb{E}_{p_v} \mathbb{E}_{p(x)} \left[v^T \nabla_x s_\theta(x)v + \frac{1}{2} \lVert s_\theta(x) \rVert^2_2 \right]
    }
\)
where \(v \sim p_v\) is a simple distribution of random vectors. \\
\textbf{Contrastive divergence.}
\citet{hintonTrainingProductsExperts2002} approximates the gradient of the maximum likelihood objective by
\(
\label{eq:cd}
    \nabla_\theta p_\theta(x) = \mathbb{E}_{p_\theta(x^\prime)} \left[ \nabla_\theta E_\theta(x^\prime) \right] - \nabla_\theta E_\theta(x)
\)
Following recent literature \cite{duImplicitGenerationGeneralization2020b}, we approximate the expectation with samples obtained through Stochastic Gradient Langevin Dynamics \cite{wellingBayesianLearningStochastic2011}. \\ % For discrete data, we use the Gibbs-with-Gradients sampler proposed in \cite{grathwohlOopsTookGradient2021}. 
\textbf{VERA.}
Lastly, we consider the recently proposed Variational Entropy Regularized Approximate maximum likelihood (VERA) training \cite{grathwohlNoMCMCMe2020} which learns the parameters \(\phi\) of a auxiliary distribution $q_\phi$ as the optimum of
\(
    \log Z(\theta) = \max_{q_\phi} \mathbb{E}_{q_\phi(x)} \left[ f_\theta(x) \right] + H(q_\phi)
\)
which can be plugged into \Cref{eq:ebm} to obtain an alternative method for training EBMs. 
%The generator distribution is chosen to be of the form \(q_\phi(x) = \int q_\phi (x \mid z) q(z) dz \). 
\citet{grathwohlNoMCMCMe2020} propose a variational approximation to estimate the entropy term \(H_{q_\phi}\) circumventing the need for sampling \cite{diengPrescribedGenerativeAdversarial2019, titsiasUnbiasedImplicitVariational2019}. \\
While more approaches for training EBMs exist, they either assume knowledge about a noise distribution close to the ground-truth data distribution \cite{gutmannNoisecontrastiveEstimationNew, ceylanConditionalNoiseContrastiveEstimation2018} or have shown to require prohibitive amounts of training time in our experiments \cite{gaoFlowContrastiveEstimation2020}.

\begin{table*}
    \centering
    \caption{AUC-PR for OOD detection on the respective in-distribution dataset.}
    \resizebox{.85\linewidth}{!}{%
    \setlength{\tabcolsep}{1mm}
    \begin{tabular}{lllllllllll}
\toprule
ID dataset & \multicolumn{5}{c}{CIFAR-10} & \multicolumn{3}{c}{FMNIST} &                                      \multicolumn{1}{c}{Segment} &                         \multicolumn{1}{c}{Sensorless} \\
\cmidrule(l){2-6} \cmidrule(l){7-9} \cmidrule(l){10-10} \cmidrule(l){11-11} 
OOD dataset &                                    CIFAR-100 &                            CelebA &                                         LSUN &                                          SVHN &                                      Textures &                                        KMNIST &                                        MNIST &                           NotMNIST &                                  Segment OOD &                     Sensorless OOD \\
\midrule
CE    &  \bfseries{62.76 {\footnotesize $\pm$ 1.46}} &  64.47 {\footnotesize $\pm$ 2.44} &             65.18 {\footnotesize $\pm$ 5.79} &              47.51 {\footnotesize $\pm$ 4.58} &              39.17 {\footnotesize $\pm$ 2.28} &              69.07 {\footnotesize $\pm$ 6.73} &  \bfseries{82.5 {\footnotesize $\pm$ 12.27}} &    50.9 {\footnotesize $\pm$ 6.73} &             33.35 {\footnotesize $\pm$ 1.82} &   33.02 {\footnotesize $\pm$ 1.32} \\
NF    &                                       58.34  &                 \bfseries{74.68 } &                                       62.99  &                                        31.58  &                                        50.23  &                                        62.22  &                                       49.03  &                  \bfseries{93.68 } &  \bfseries{99.45 {\footnotesize $\pm$ 0.18}} &                  \bfseries{94.35 } \\
\midrule
CD  &             50.51 {\footnotesize $\pm$ 2.13} &  43.86 {\footnotesize $\pm$ 5.85} &            54.43 {\footnotesize $\pm$ 11.37} &  \bfseries{60.72 {\footnotesize $\pm$ 24.59}} &  \bfseries{76.21 {\footnotesize $\pm$ 17.44}} &              50.52 {\footnotesize $\pm$ 9.39} &              31.69 {\footnotesize $\pm$ 0.9} &   76.85 {\footnotesize $\pm$ 2.66} &             98.18 {\footnotesize $\pm$ 2.18} &  72.83 {\footnotesize $\pm$ 16.19} \\
SSM   &             53.82 {\footnotesize $\pm$ 3.12} &   57.72 {\footnotesize $\pm$ 7.0} &             52.79 {\footnotesize $\pm$ 3.16} &              45.75 {\footnotesize $\pm$ 7.24} &              48.82 {\footnotesize $\pm$ 4.34} &              58.98 {\footnotesize $\pm$ 5.48} &             67.86 {\footnotesize $\pm$ 11.4} &  57.27 {\footnotesize $\pm$ 13.73} &            79.43 {\footnotesize $\pm$ 24.29} &  67.14 {\footnotesize $\pm$ 20.31} \\
VERA  &             55.95 {\footnotesize $\pm$ 2.68} &  73.97 {\footnotesize $\pm$ 2.63} &  \bfseries{67.39 {\footnotesize $\pm$ 2.57}} &              37.27 {\footnotesize $\pm$ 4.66} &               46.29 {\footnotesize $\pm$ 8.1} &  \bfseries{78.11 {\footnotesize $\pm$ 21.05}} &            67.53 {\footnotesize $\pm$ 21.63} &  76.22 {\footnotesize $\pm$ 22.11} &             94.63 {\footnotesize $\pm$ 7.22} &  45.66 {\footnotesize $\pm$ 10.55} \\
\bottomrule
\end{tabular}
    }
    \resizebox{.85\linewidth}{!}{%
    \begin{tabular}{lllllllllll}
\toprule
ID dataset & \multicolumn{3}{c}{CIFAR-10} & \multicolumn{3}{c}{FMNIST} & \multicolumn{2}{c}{Segment} & \multicolumn{2}{c}{Sensorless} \\
\cmidrule(l){2-4} \cmidrule(l){5-7} \cmidrule(l){8-9} \cmidrule(l){10-11} 
OOD dataset &                                      Constant &                                       Noise &                                     OODomain &                           Constant &                                       Noise &                                    OODomain &                                    Constant &                                       Noise &                                    Constant &                                       Noise \\
\midrule
CE    &               45.26 {\footnotesize $\pm$ 8.8} &           61.13 {\footnotesize $\pm$ 21.02} &              30.69 {\footnotesize $\pm$ 0.0} &    35.5 {\footnotesize $\pm$ 3.08} &           55.84 {\footnotesize $\pm$ 22.32} &            30.74 {\footnotesize $\pm$ 0.11} &           41.74 {\footnotesize $\pm$ 18.57} &            33.66 {\footnotesize $\pm$ 2.77} &            32.38 {\footnotesize $\pm$ 1.19} &            31.97 {\footnotesize $\pm$ 1.26} \\
NF    &                                        30.87  &            83.65 &                                            \bfseries{100.0} &                  \bfseries{71.07 } &            98.04 &                                           \bfseries{100.0} &            99.95  &  \bfseries{100.0} &                           \bfseries{100.0 } &  \bfseries{100.0} \\
\midrule
CD  &  \bfseries{58.75 {\footnotesize $\pm$ 28.17}} &  \bfseries{100.0 {\footnotesize $\pm$ 0.0}} &            58.41 {\footnotesize $\pm$ 37.96} &  70.59 {\footnotesize $\pm$ 12.84} &  \bfseries{100.0 {\footnotesize $\pm$ 0.0}} &  \bfseries{100.0 {\footnotesize $\pm$ 0.0}} &            95.47 {\footnotesize $\pm$ 2.34} &            95.14 {\footnotesize $\pm$ 3.71} &  \bfseries{100.0 {\footnotesize $\pm$ 0.0}} &  \bfseries{100.0 {\footnotesize $\pm$ 0.0}} \\
SSM   &             47.24 {\footnotesize $\pm$ 15.56} &           70.28 {\footnotesize $\pm$ 31.39} &  68.57 {\footnotesize $\pm$ 25.2} &  47.57 {\footnotesize $\pm$ 15.18} &           49.45 {\footnotesize $\pm$ 21.19} &            76.76 {\footnotesize $\pm$ 21.1} &           73.91 {\footnotesize $\pm$ 25.44} &            81.1 {\footnotesize $\pm$ 17.87} &            69.79 {\footnotesize $\pm$ 6.62} &           64.61 {\footnotesize $\pm$ 17.35} \\
VERA  &              31.51 {\footnotesize $\pm$ 0.66} &  \bfseries{100.0 {\footnotesize $\pm$ 0.0}} &            63.48 {\footnotesize $\pm$ 34.37} &  53.24 {\footnotesize $\pm$ 22.65} &           79.34 {\footnotesize $\pm$ 27.34} &           72.42 {\footnotesize $\pm$ 37.61} &  \bfseries{100.0 {\footnotesize $\pm$ 0.0}} &  \bfseries{100.0 {\footnotesize $\pm$ 0.0}} &  \bfseries{100.0 {\footnotesize $\pm$ 0.0}} &            99.85 {\footnotesize $\pm$ 0.31} \\
\bottomrule
\end{tabular}
    }
    \label{tab:all_results}
\end{table*}

\section{Experiments}
\label{ch:experiments}
We investigate the OOD detection performance of EBMs trained with the approaches discussed in \Cref{ch:method}.
%
% \todo{Small discussion of semantic/high-level/discriminative features vs. low-level/non-semantic features with example, make clear that we use these as synonyms}
%
In particular, we verify the following hypotheses improving OOD detection with EBMs in recent works \cite{grathwohlYourClassifierSecretly2020, grathwohlNoMCMCMe2020} compared to Normalizing Flows: \\
\textbf{Dimensionality reduction.} The manifold hypotheses \cite{feffermanTestingManifoldHypothesis2013} suggests that high-dimensional data such as images reside on a lower-dimensional manifold. Normalizing Flows require invertible transformations and thus operate in the original data space. We hypothesize that this hinders OOD detection as they need to model off-manifold directions. Contrarily, EBMs do not require invertibility, which allows pruning of redundant dimensions without semantic content. \\ % \bc{I felt the last sentence not convincing. We could say that EBM can do dim reduction, thus hopefully removing useless dimensions}
\textbf{Supervision.} \citet{kirichenkoWhyNormalizingFlows2020, schirrmeisterUnderstandingAnomalyDetection2020} show that Normalizing Flows learn low-level features without semantic meaning (smoothness, etc.) common to all natural images \cite{serraInputComplexityOutofdistribution2020}. We hypothesize that label information encourages semantic, high-level features instead, improving OOD detection.

\textbf{Setup.}
We perform OOD detection by comparing the density of ID and OOD inputs under the learned \(p_\theta(x)\). For evaluation, we consider OOD detection as a binary classification problem with labels \(1\) for ID and \(0\) for OOD and report average precision (AUC-PR) as commonly done in the literature \cite{hendrycksBaselineDetectingMisclassified2018}. We train EBMs with Sliced Score Matching (\textit{SSM}), Contrastive Divergence (\textit{CD}), and \textit{VERA} as described in \Cref{ch:method}. For baselines, we compare with Normalizing Flow (\textit{NF}) and the energy score of a classifier \cite{liuEnergybasedOutofdistributionDetection2020} (\textit{CE}) \footnote{We provide code at \url{https://github.com/selflein/EBM-OOD-Detection}}. \\
\textbf{Datasets.}
Following \cite{charpentierPosteriorNetworkUncertainty2020}, we consider the tabular datasets \textit{Sensorless drive} and \textit{Segment}, with dimensionality 18 and 49 and 4 and 11 classes, respectively. To obtain a representative OOD dataset, we remove one class (\textit{sky}) from Segment and two classes (\textit{10}, \textit{11}) from Sensorless drive.
%
% Further, we use the bacteria genomics dataset proposed in \cite{xiaoLikelihoodRegretOutofDistribution2020}. The data consists of sequences of length 250 encoding the type of nucleobases at each position together with a label corresponding to the bacteria class. The dataset consists of a in-distribution set of 10 bacteria classes discovered before 2011, an OOD validation set with 60 classes discovered between 2011 and 2015, and a OOD test set with additional 60 classes discovered after 2015.
%
Further, we evaluate on image datasets. We use FMNIST \cite{xiaoFashionMNISTNovelImage2017} as ID dataset and MNIST \cite{lecunGradientbasedLearningApplied1998}, NotMNIST \cite{bulatovMachineLearningEtc2011}, KMNIST \cite{clanuwatDeepLearningClassical2018} as OOD datasets. Additionally, we train on CIFAR-10 \cite{krizhevskyLearningMultipleLayers2009} and use LSUN \cite{yuLSUNConstructionLargescale2016}, Textures \cite{huangCompactConvolutionalNeural2020}, CIFAR-100 \cite{krizhevskyLearningMultipleLayers2009}, SVHN \cite{netzerReadingDigitsNatural2011} and Celeb-A \cite{liu2015faceattributes} as out-of-distribution. In the following, we refer to these OOD datasets as \textit{natural} OOD datasets.
Finally, we generate \textit{non-natural} OOD datasets with \textit{noise} and \textit{constant} input. As proposed by \citet{charpentierPosteriorNetworkUncertainty2020}, we also consider an \textit{OODomain} dataset where the input data is not normalized into the range \( [0, 1] \).\\
\underline{\textit{Natural vs. non-natural datasets.}} Note that the differentiation of \textit{natural} and \textit{non-natural} datasets allows evaluating distinct properties of the learned density: A model able to distinguish \textit{natural} inputs can recognize semantic features of the high-level content of images, e.g., corresponding to classes, while \textit{non-natural} inputs are easily detected semantically but lie farther away from the data manifold, thus, require the model to decrease the density when moving away from the data distribution.
%Distinguishing \textit{natural} inputs requires learning semantic features regarding the high-level content of images, while {non-natural} inputs are easily detected semantically but lie further outside the data manifold, thus, requiring a model where the density decreases when moving away from the data distribution. \bc{I feel taht this distinction between antural and non-natural data is interesting. Would it be possible elaborate more about the motivations for this distinction and what we expect to learn from each type of OOD ?}
%
\textbf{Architectures.}
We use MLPs on the tabular datasets
%, a 1-D convolutional network for the genomics dataset as in \cite{renLikelihoodRatiosOutofDistribution2019}
and WideResNet-10-2 \cite{zagoruykoWideResidualNetworks2017} for the image datasets. For the Normalizing Flow baseline, radial flows \cite{rezendeVariationalInferenceNormalizing2016} are used on the tabular and Glow \cite{kingmaGlowGenerativeFlow2018} on the image datasets. We provide more details in \Cref{sec:architecture_details}.

% Group w.r.t. dataset type (Tabular datasets, Sequential dataset, image datasets)
% Group w.r.t. experiment type (vanilla EBM, EBM with supervision, EBM on embeddings, Adjusted architecture)

% \paragraph{Experiments}
% \begin{itemize}
%     \item Baseline general results (comparison to Normalizing Flows/ Cross-Entropy baseline)
%     \item Additional Supervision
%     \begin{itemize}
%         \item improves results in cases and reduces performance in others (show relative improvements per dataset averaged over OOD datasets?)
%         \item Weighting of classification objective matters and requires tuning. 
%     \end{itemize}
%     \item Train on embeddings obtained from Cross Entropy classifier (-> Show relative improvements)
%     \item Architectural changes to enforce higher level features (-> Show relative improvements)
% \end{itemize}

% \bc{For each experiment, I think it would be nice to organize them explicitly like this: \\
% (1) what is evaluation setup, \\
% (2) what we observe, and \\
% (3) what we conlude on ech dataset type. I feel that then it is easier for the reader to find the information}

\subsection{Are EBMs better than baselines in general?}
\textbf{Experiment 1.}
We establish baseline results by training EBMs and baseline models.
%
% \bc{The training and evaluation metrics still sounds like the setup to me. Could we move it the previous section ?}
In \Cref{tab:all_results}, we find that EBMs consistenly outperform the \textit{CE} baseline by \(62.9\%\), \(55.0\%\), and \(36.4\%\) for \textit{CD}, \textit{VERA}, and \textit{SSM} respectively. The improvements are moderate in comparison to the Normalizing Flow baselines with \(11.9\%\), \(4.3\%\), and \(-4.3\%\). Notably, improvements are mostly on \textit{natural} datasets. % \bc{This is already a first conclusion, right ? It answer the question, "are EBMs better than other baselines in general ?"}
As EBMs perform dimensionality reduction since they map
%the EBM is specified by a function $\mathbb{R}^D \mapsto \mathbb{R}$ mapping
to the scalar energy and do not consistently outperform Normalizing Flows in this experiment across all training methods, we conclude that dimensionality reduction plays a minor role in the OOD detection performance of recent EBMs \cite{grathwohlYourClassifierSecretly2020}. We attribute slight improvements on \textit{natural} data to the ability to discard non-semantic dimensions in EBMs.
% \begin{table*}
%     \centering
%     \caption{AUC-PR for OOD detection on non-natural datasets for models trained with the EBM training approaches on the in-distribution dataset.}
%     \resizebox{\linewidth}{!}{%
%     \input{tables/overall_table_other}
%     }
%     \label{tab:all_results_fake}
% \end{table*}
%
\begin{figure*}
    \centering
    \begin{minipage}[t]{.7\linewidth}
         \includegraphics[width=\linewidth]{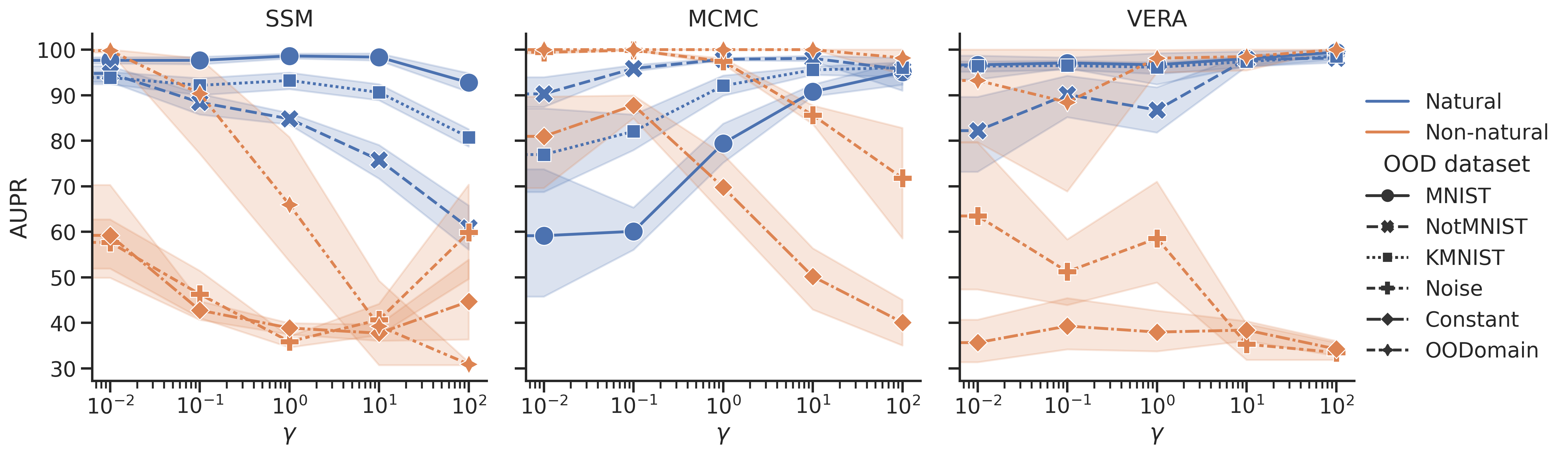}
         \vspace{-8mm}
         \caption{AUC-PR for OOD detection for different settings of the weighting hyperparameter \(\gamma\) of the cross entropy objective. FMNIST is used as the in-distribution dataset.}
         \label{fig:fashionmnist_clf_weight}
    \end{minipage}%
    \hfill
    \begin{minipage}[t]{.28\linewidth}
         \includegraphics[width=\linewidth]{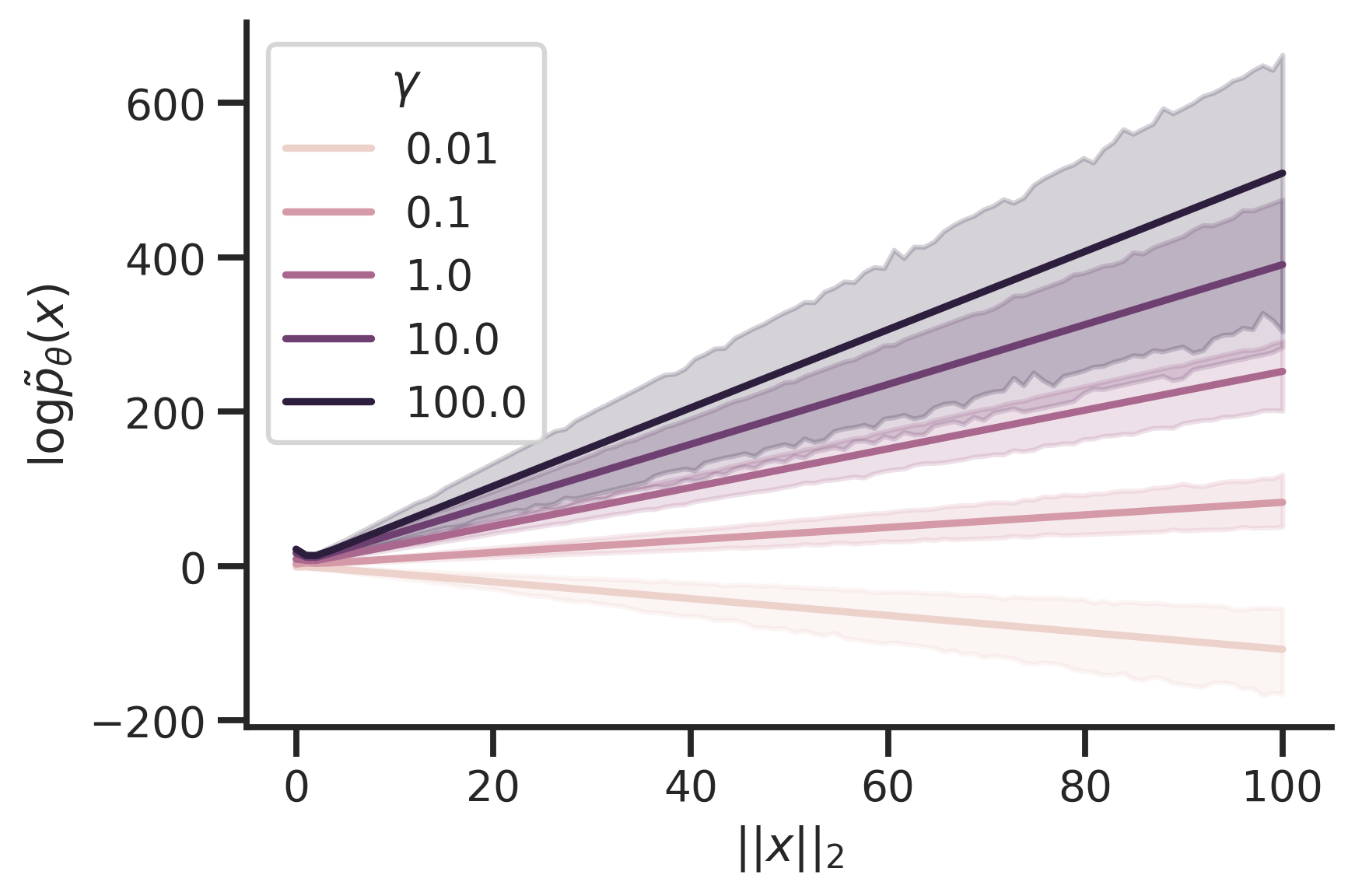}
         \vspace{-8mm}
         \caption{Unnormalized density \(\tilde{p}_\theta(x)\) of inputs with increasing $L_2$-norm.% esimated with models trained with SSM and supervision with different \(\gamma\).
         }
         \label{fig:ssm_moving_away_from_training_data}
    \end{minipage}
\end{figure*}
\subsection{Does supervision improve OOD detection?}
Next, we consider two ways of incorporating labels to investigate the influence of supervision. Firstly, by applying an additional loss term as in \cite{grathwohlYourClassifierSecretly2020} which affects the optimization directly, and secondly, performing density estimation on embeddings of a classification model which incorporates supervision indirectly through class-related features.\\
\textbf{Experiment 2.}
We consider JEMs as introduced in \Cref{ch:method} and apply a cross entropy objective with weighting hyperparameter \(\gamma\) optimizing \(p_\theta(y \mid x)\).
In \Cref{tab:supervision_improvement}, we find substantial improvements in OOD detection on most datasets compared to the baseline models.
Using label information within the model encourages discriminative features relevant for classification, improving detection of \textit{natural}, OOD inputs by $29.61\%$. These results indicate that EBM training tends to assign high-likelihood to all natural, structured images, an issue also observed in other generative models \cite{renLikelihoodRatiosOutofDistribution2019}. Note however that supervision decreases performance on some \textit{natural} datasets and consistently worsens results at differentiating \textit{non-natural} inputs ($-11.74\%$). 
%
% \paragraph{Experiment 3}
% In order to identify the cause of the mismatch between the OOD detection on \textit{natural} and \textit{non-natural} datasets, we investigate the % factorization of the joint distribution \(p_\theta(x, y) = p_\theta(x) + \gamma p_\theta(y \mid x)\) used for training.
% In particular, we investigate the influence of the weighting parameter \(\gamma\) by training EBMs with different settings of this hyperparameter.
%
\begin{table}
    \centering
    \caption{\% improvement in AUC-PR for OOD detection when using additional supervision during training.}
    \label{tab:supervision_improvement}
    \resizebox{.3\textwidth}{!}{
    \begin{tabular}{llrr}
\toprule
 Model    &ID dataset &  Natural &  Non-natural \\
\midrule
\multirow{4}{*}{CD} & CIFAR-10 &   -10.82 &      -9.11 \\
     & FMNIST &    47.17 &       3.24 \\
     & Segment &     1.85 &       0.89 \\
     & Sensorless &    29.72 &      -0.02 \\
\midrule
\multirow{4}{*}{SSM} & CIFAR-10 &     7.33 &     -27.94 \\
     & FMNIST &    50.61 &     -20.26 \\
     & Segment &    25.89 &     -21.94 \\
     & Sensorless &    22.13 &     -40.73 \\
\midrule
\multirow{4}{*}{VERA} & CIFAR-10 &    -1.16 &      -3.00 \\
     & FMNIST &    33.66 &     -15.53 \\
     & Segment &     4.98 &      -0.57 \\
     & Sensorless &    97.93 &       0.07 \\
\bottomrule
\end{tabular}

    }
\end{table}
Investigating the difference in results between \textit{natural} and \textit{non-natural} datasets, we observe in \Cref{fig:fashionmnist_clf_weight} that the OOD detection on \textit{non-natural} images is negatively impacted by increasing weighting of the cross-entropy objective. In \Cref{fig:ssm_moving_away_from_training_data}, we observe that the EBM assigns exponentially increasing density to datapoints distant from the training data distribution for higher settings of \(\gamma\) similar to what has been proven for the confidence in ReLU networks \cite{heinWhyReLUNetworks2019}. 
% In \Cref{thm:jem_unbounded_energy}, we show that this relation also holds for the density in a JEM trained with cross-entropy objective supporting our experimental results.
%
As a result \textit{non-natural} inputs which are further away from the training data than \textit{natural} images become increasingly harder to detect. 
%Further note that as the density increases unbounded, the learned energy function probably \textbf{does not describe a valid distribution at all} \bc{I agree that this statement is quite interesting. I think that it deserves to explain why in practice this might happen while theoretically/mathematically this should work.}.
We conclude that training with this factorization requires tuning of \(\gamma\) to achieve high OOD detection performance on both \textit{natural} and \textit{non-natural} inputs. \\
% \bc{Maybe give a reference to "Why Relu networks are overconfident" paper}
% Additionally, we observe that setting \(\gamma\) too high degrades performance detecting the \textit{natural} OOD datasets in some cases.   
%
%
\textbf{Experiment 3.}
Sidestepping the issue of tuning \(\gamma\), we follow \citet{kirichenkoWhyNormalizingFlows2020} noticing that training Normalizing Flows on high-level features improves OOD detection. 
% As EBMs do not require architectural restrictions, they should be able to extract those semantic features without additional processing. 
To investigate this behavior for EBMs, we store the features from a classifier trained with cross-entropy objective after convolutional layers. Subsequently, we train EBMs on these embeddings.
% Note that the dimensionality is not reduced significantly. For FMNIST, we obtain embeddings of size \(640\) while the original dimensionality is only \(28 \times 28 = 784\). 
%
In \Cref{tab:embedding_improvemnt}, we observe that density estimation on embeddings significantly improves results on \textit{natural} datasets compared to the baseline trained on images directly ($+53.65\%$). Further, performance on \textit{non-natural} datasets does not deteriorate with this approach and increases performance by $10.98\%$ on average.
As training on discriminative features directly improves OOD detection, this supports our hypotheses that EBMs trained on high-dimensional data such as images struggle to learn semantic features. 
%Overall, this supports that EBMs assign high-likelihood to all (even OOD), structured images if not explicitly encouraged to learn semantic features. 
%\bc{This conclusion is not clear to me: they cannot learn features to detect OOD but assign them lower likelihood ?}\bc{In tab 1, it looks like EBMs still fail in some cases without supervision, right ?}
%
\begin{table}
    \centering
    \setlength{\tabcolsep}{2pt}
    \begin{minipage}[t]{.48\linewidth}
        \caption{\% improvement in AUC-PR for OOD detection when training on embeddings.}
        \label{tab:embedding_improvemnt}
        \resizebox{\textwidth}{!}{
        \begin{tabular}{llrr}
\toprule
 Model    & ID dataset &  Natural & Non-natural \\
\midrule
\multirow{2}{*}{CD} & CIFAR-10 &    48.60 &       3.37 \\
     & FMNIST &    95.79 &     -13.52 \\
\midrule
\multirow{2}{*}{SSM} & CIFAR-10 &    53.84 &      -2.31 \\
     & FMNIST &    58.40 &      59.59 \\
\midrule
\multirow{2}{*}{VERA} & CIFAR-10 &    50.16 &      16.97 \\
     & FMNIST &    15.12 &       1.80 \\
\bottomrule
\end{tabular}

        }
    \end{minipage}%
    \hfill
    \begin{minipage}[t]{.48\linewidth}
        \caption{\% improvement in AUC-PR for OOD detection after introducing bottlenecks.}
        \label{tab:dim_red_improvement}
        \resizebox{\textwidth}{!}{
        \begin{tabular}{llrr}
\toprule
 Model    & ID dataset &  Natural &  Non-natural \\
\midrule
\multirow{2}{*}{CD} & CIFAR-10 &    20.18 &        20.38 \\
     & FMNIST &    67.95 &        10.88 \\
\midrule
\multirow{2}{*}{SSM} & CIFAR-10 &    14.76 &        33.34 \\
     & FMNIST &     1.75 &        -5.92 \\
\midrule
\multirow{2}{*}{VERA} & CIFAR-10 &    19.66 &        33.22 \\
     & FMNIST &    26.84 &        32.94 \\
\bottomrule
\end{tabular}

        }
    \end{minipage}
\end{table}
\subsection{Can we encourage semantic features?}
While EBMs inherently perform dimensionality reduction, the previous experiments suggest this being insufficient to capture semantic features within the data. 
As shown by \citet{kirichenkoWhyNormalizingFlows2020}, introducing a bottleneck in the coupling transforms of Normalizing Flows enforces the network to learn semantic features improving OOD detection. This can also be interpret in the frame of compression \cite{serraInputComplexityOutofdistribution2020} where redundant information is removed.\\
\textbf{Experiment 4.}
We introduce bottlenecks after every block of the WRN through a set of $\smash{1\times1}$ convolutions mapping to $0.2 \times$ the original dimensionality. %We experiment with different values of the downscaling factor and $0.2$ did yield the best results, however, other settings did improve OOD detection as well. Full results can be found in the appendix.
In \Cref{sec:additional_results}, we provide results for other settings of the bottleneck.
In \Cref{tab:dim_red_improvement}, we observe that this simple adjustment yields improvements in OOD detection on \textit{natural} images for all training methods. 
The bottlenecks force the network to compress the features removing redundant information and enable improved OOD detection supporting the hypotheses that generic EBMs retain non-semantic features. We provide further investigation on low-level features in \Cref{sec:additional_results:low_level_features}.
% \bc{It might be worth to be more precise for the conclusion sentence: what is the link between high level features and low dimensions ?}

\section{Conclusion}
\label{ch:conclusion}

% \paragraph{Next steps}
% \begin{itemize}
%     \item Optimize joint $p(x, y)$ instead of factorization which we show to require per dataset tuning
%     \item Incorporate low-dim. semantic latent space (latent space EBMs)
% \end{itemize}
% In this work, we introduce and find supporting evidence for the hypotheses that supervision and dimensionality reduction facilitate improved OOD detection in EBMs.
% Our adjustments isolating each of the hypotheses enable higher-level, semantic features which we find to be relevant for distinguishing OOD inputs not only on images but also on higher-dimensional tabular datasets.
%This provides insights for future work aiming to improve the OOD detection in EBMs. Firstly, incorporating supervision and, secondly, incorporating architectural restrictions which enable higher level features. Further, we have seen that optimizing the factorization of the joint distribution \(p(x, y)\) proposed in the literature exhibits issues concerning additional hyperparameters. As a result a further direction could be optimizing the joint distribution directly circumventing the hybrid training approach.

Overall, we find that \textbf{(1)} EBMs struggle with OOD detection on high-dimensional data but to a lower degree than Normalizing Flows, \textbf{(2)} incorporating task-specific priors such as supervision significantly improves OOD detection on \textit{natural} OOD data% in line with what has been recently suggested by \citet{lanPerfectDensityModels2021, schirrmeisterUnderstandingAnomalyDetection2020} for other generative models
, and \textbf{(3)} architectural modifications can be used to improve the OOD detection performance. 

% This provides insights for future work aiming to improve the OOD detection in EBMs. Firstly, incorporating supervision and, secondly, incorporating architectural restrictions which enable higher level features. Further, we have seen that optimizing the factorization of the joint distribution \(p(x, y)\) proposed in the literature exhibits issues concerning additional hyperparameters. As a result a further direction could be optimizing the joint distribution directly circumventing the hybrid training approach.

% Future work could aim at investigating which and why some architectural modifications yield improved OOD detection. Further, considering other factorizations such as the joint \(p_\theta(x, y)\) for training JEMs \citet{grathwohlYourClassifierSecretly2020} could be a compelling direction since we found that the current method can lead to degenerate solutions.

\bibliography{paper}
\bibliographystyle{icml2021}

%\appendix
%\section{Do \emph{not} have an appendix here}
%
%\textbf{\emph{Do not put content after the references.}}
%%
%Put anything that you might normally include after the references in a separate
%supplementary file.
%
%We recommend that you build supplementary material in a separate document.
%If you must create one PDF and cut it up, please be careful to use a tool that
%doesn't alter the margins, and that doesn't aggressively rewrite the PDF file.
%pdftk usually works fine. 
%
%\textbf{Please do not use Apple's preview to cut off supplementary material.} In
%previous years it has altered margins, and created headaches at the camera-ready
%stage. 
%%%%%%%%%%%%%%%%%%%%%%%%%%%%%%%%%%%%%%%%%%%%%%%%%%%%%%%%%%%%%%%%%%%%%%%%%%%%%%%
%%%%%%%%%%%%%%%%%%%%%%%%%%%%%%%%%%%%%%%%%%%%%%%%%%%%%%%%%%%%%%%%%%%%%%%%%%%%%%%

\clearpage
\appendix

\addtolength{\textfloatsep}{8mm}
\addtolength{\dbltextfloatsep}{4mm}
\addtolength{\floatsep}{4mm}

\icmltitle{Appendix}

\section{Additional results}
\label{sec:additional_results}
For completeness, we report the full results for OOD detection on \textit{natural} datasets in \Cref{tab:full_overall_table} and on \textit{non-natural} datasets in \Cref{tab:full_overall_table_other}. Model with \textit{-E} suffix correspond to models trained on embeddings of the classifier, while models with the \textit{-S} suffix correspond to model trained with additional supervision in the form of cross-entropy objective weighted with parameter $\gamma=1$.

We also present the results for different choices of the bottleneck dimensionality in \Cref{tab:bottleneck_full}.

\begin{figure*}
    \centering
    \includegraphics[width=\linewidth]{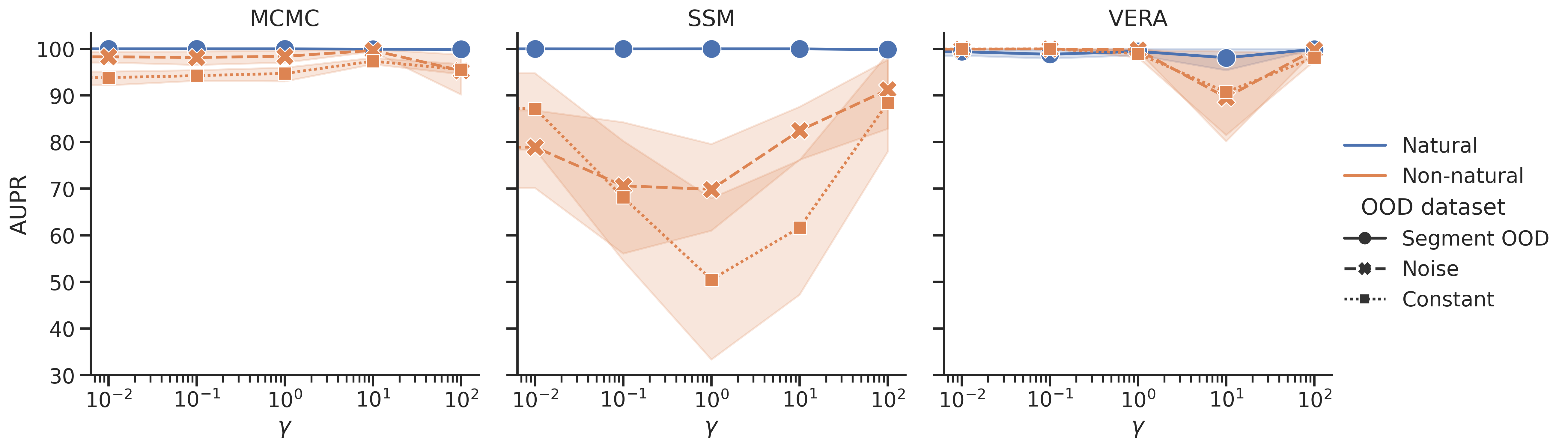}
    \caption{AUC-PR for OOD detection for different settings of the weighting hyperparameter \(\gamma\) of the cross entropy objective. Segment is used as the in-distribution dataset.}
    \label{fig:segment_clf_weight}
\end{figure*}

\begin{figure*}
    \centering
    \includegraphics[width=\linewidth]{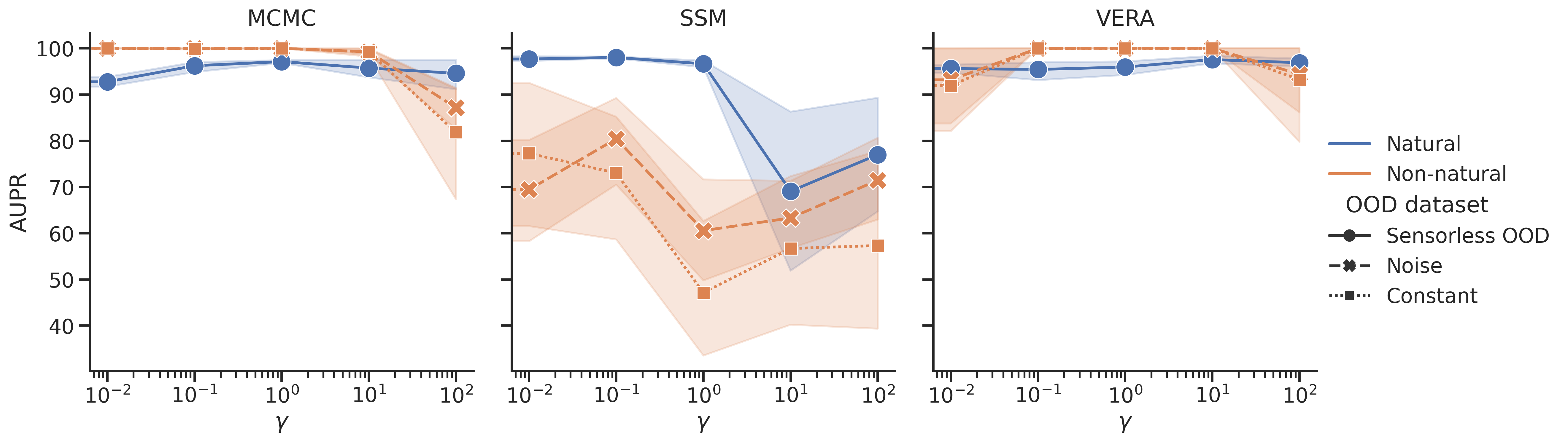}
    \caption{AUC-PR for OOD detection for different settings of the weighting hyperparameter \(\gamma\) of the cross entropy objective. Sensorless is used as the in-distribution dataset.}
    \label{fig:sensorless_clf_weight}
\end{figure*}

\begin{figure*}
    \centering
    \includegraphics[width=\linewidth]{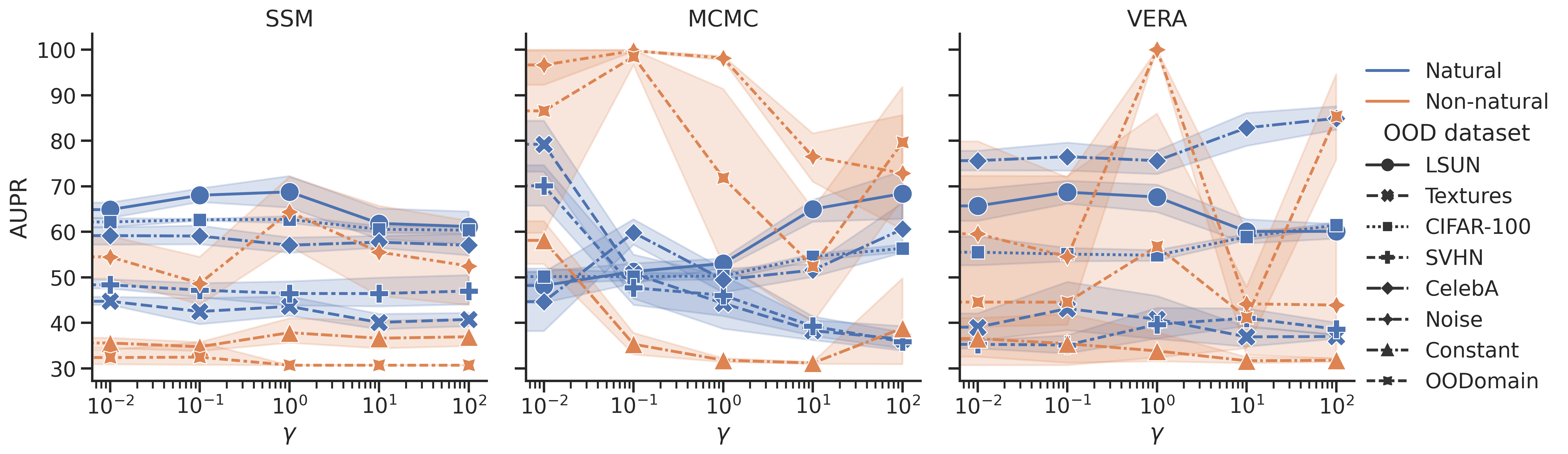}
    \caption{AUC-PR for OOD detection for different settings of the weighting hyperparameter \(\gamma\) of the cross entropy objective. CIFAR10 is used as the in-distribution dataset.}
    \label{fig:cifar10_clf_weight}
\end{figure*}

In addition to the results on the effect of the weighting parameter \(\gamma\) of the cross-entropy loss on OOD detection in EBMs on FMNIST in the main paper, we add results for the Segment dataset in \Cref{fig:segment_clf_weight}, the Sensorless dataset in \Cref{fig:sensorless_clf_weight} and CIFAR-10 in \Cref{fig:cifar10_clf_weight}. Our findings hold that the choice of \(\gamma\) heavily affects the OOD detection performance in particular on high-dimensional datasets.

\subsection{Low-level features in EBMs}
\label{sec:additional_results:low_level_features}
In the main paper, we argue that supervision encourages semantic features while unsupervised EBMs learn generic local pixel correlations (low-level features) common to all \textit{natural} images as shown by \citet{schirrmeisterUnderstandingAnomalyDetection2020} which results in worse OOD detection performance on these datasets.

\begin{figure}[h]
    \centering
    \includegraphics[width=0.2\linewidth]{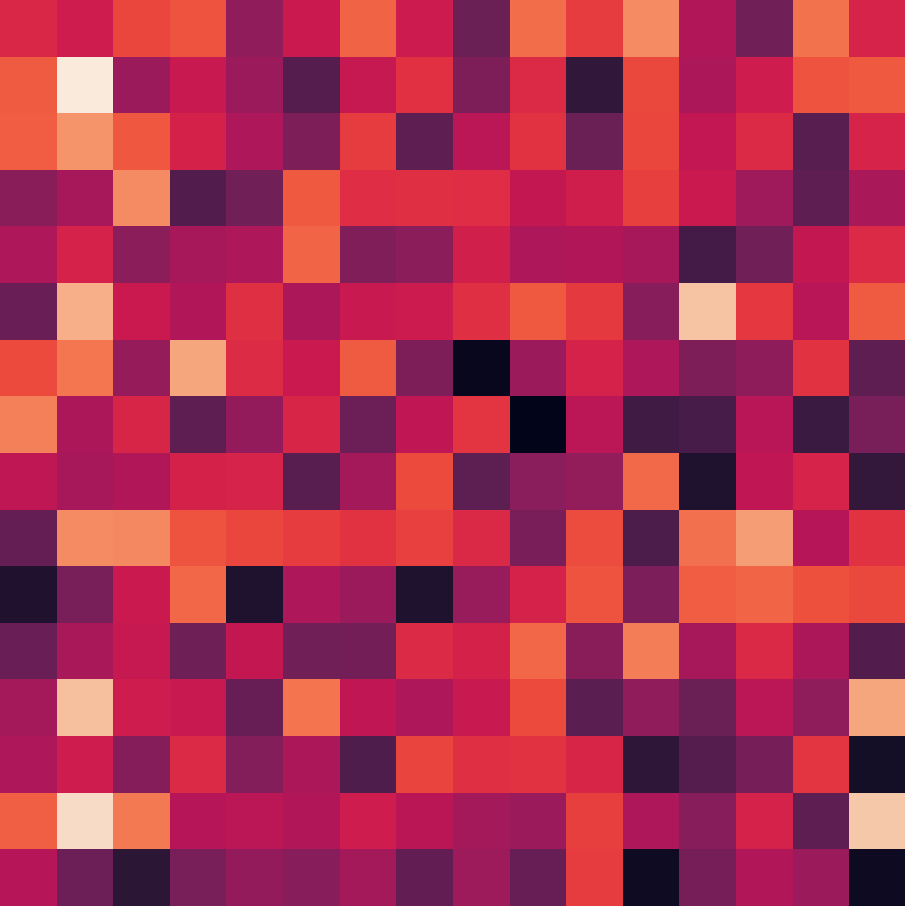}\hfill
    \includegraphics[width=0.2\linewidth]{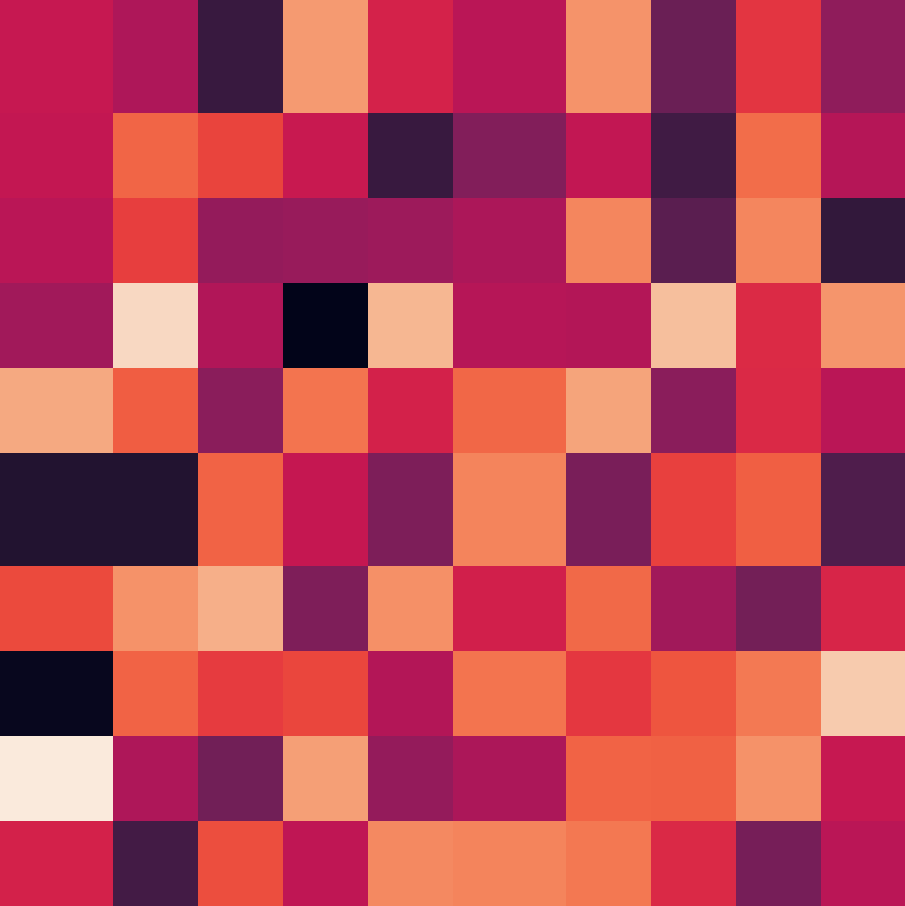}\hfill
    \includegraphics[width=0.2\linewidth]{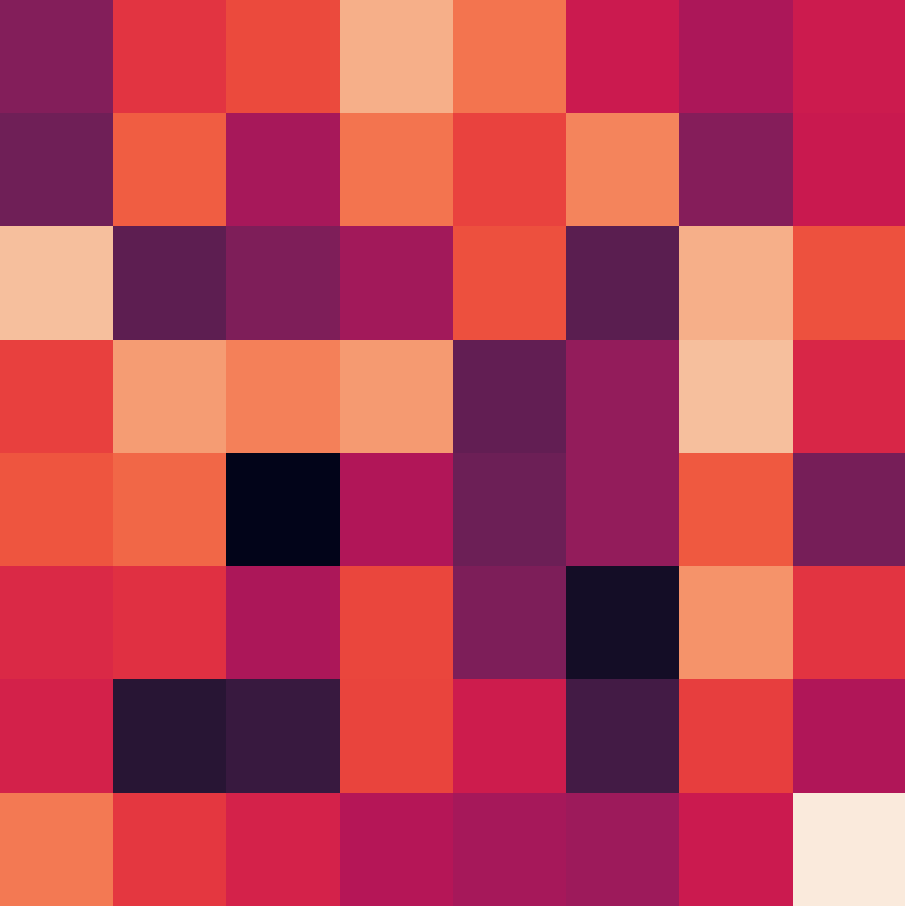}\hfill
    \includegraphics[width=0.2\linewidth]{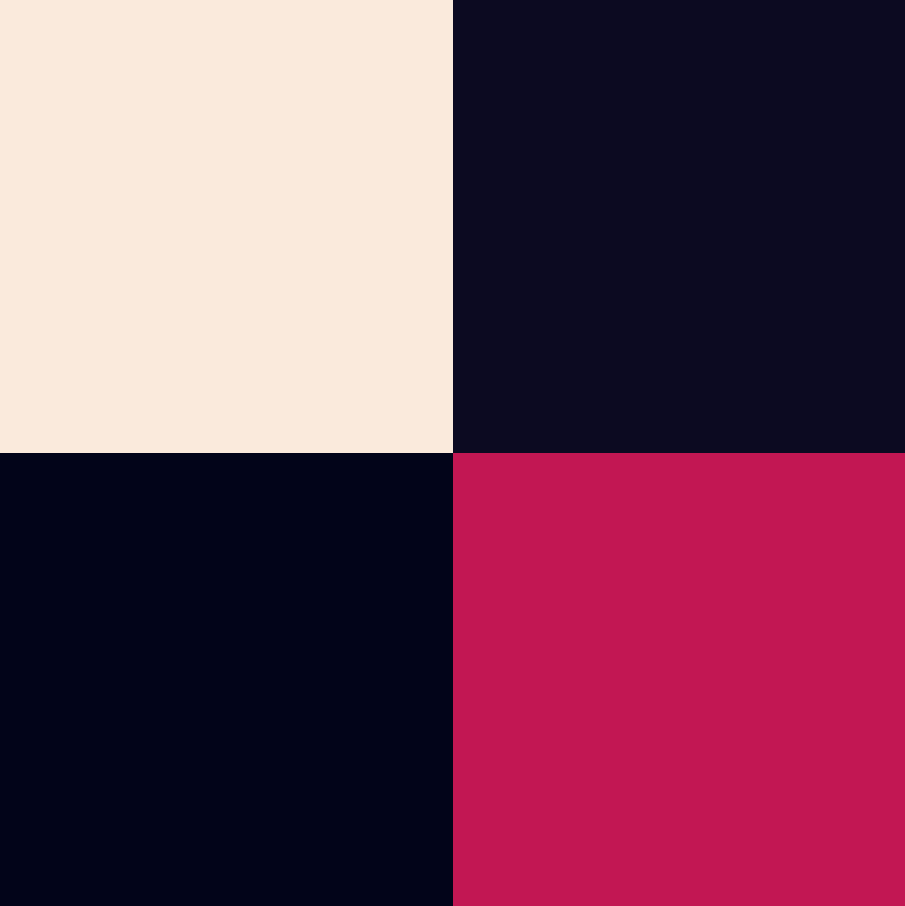}
    \caption{Example images generated with pooling sizes 2, 3, 4, and 16. Note that images become smoother the higher the pooling size.}
    \label{fig:example_input_images_smooth}
\end{figure}

\begin{figure*}
    \centering
    \begin{subfigure}[t]{0.45\linewidth}
        \centering
        \includegraphics[width=0.8\linewidth]{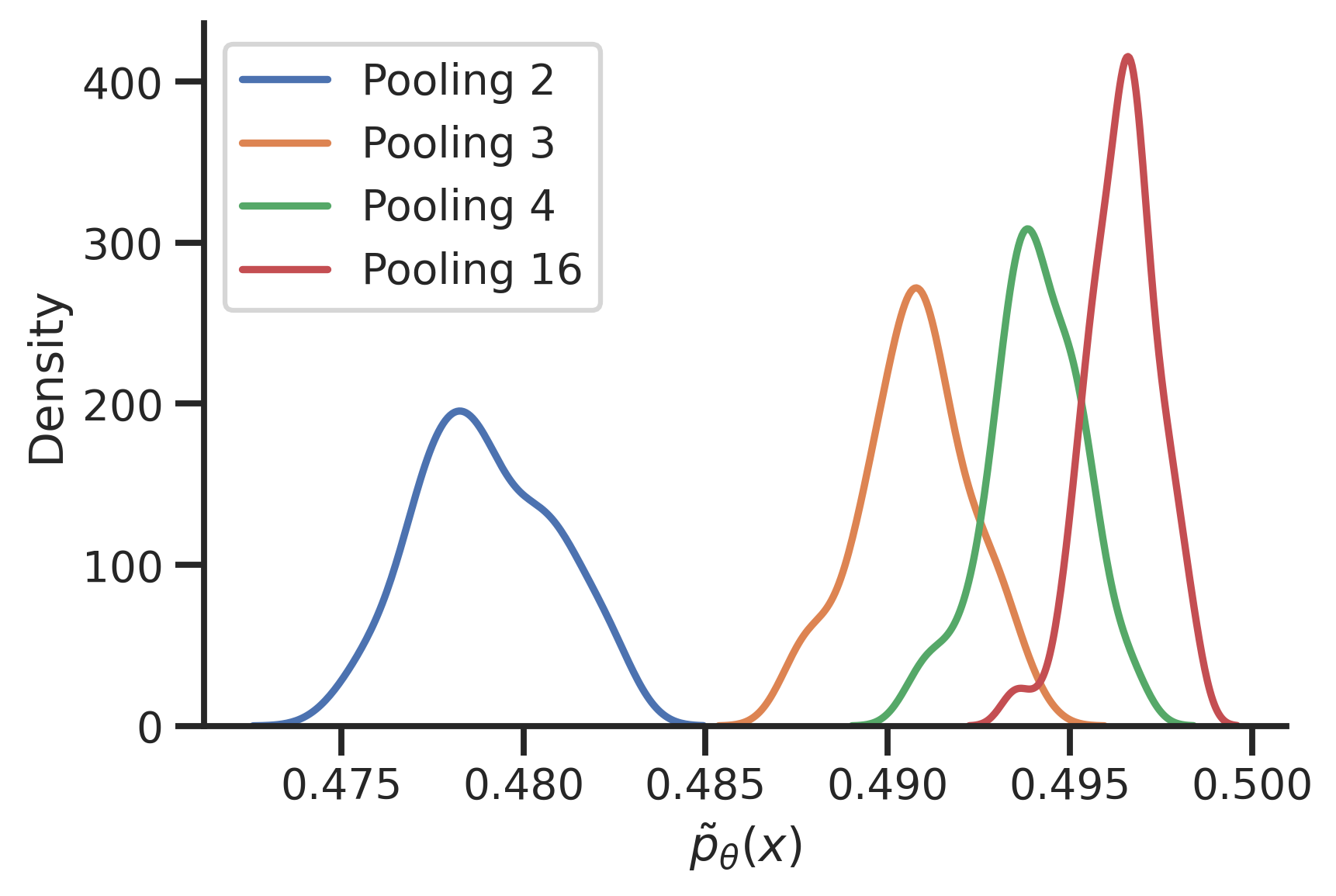}
        \caption{Unsupervised EBM.}
        \label{fig:density_histogram_smoothness_unsupervised}
    \end{subfigure}%
    \begin{subfigure}[t]{0.45\linewidth}
        \centering
        \includegraphics[width=0.8\linewidth]{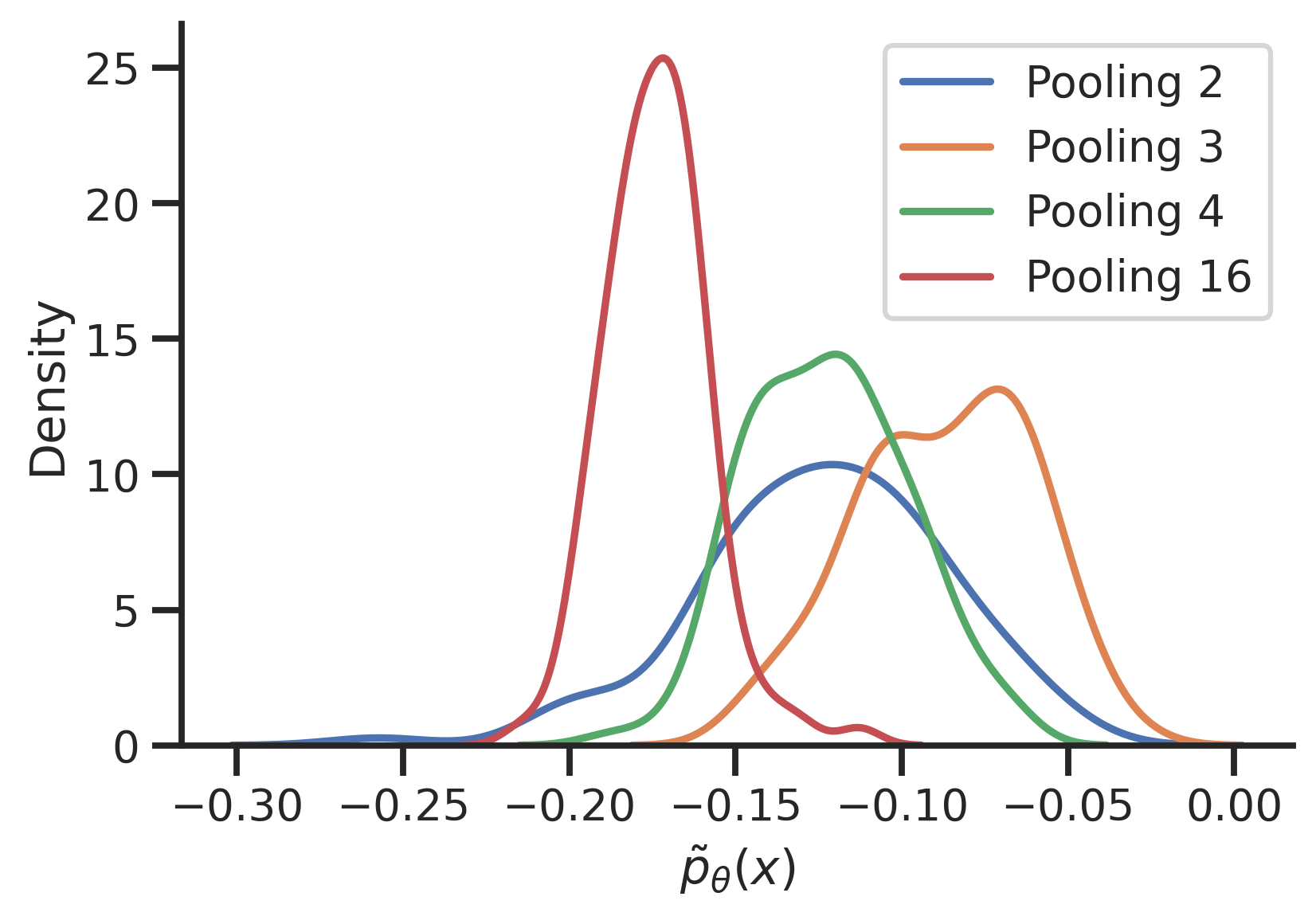}
        \caption{Supervised EBM.}
        \label{fig:density_histogram_smoothness_supervised}
    \end{subfigure}%
    \\
    \begin{subfigure}[t]{0.45\linewidth}
        \centering
        \includegraphics[width=0.8\linewidth]{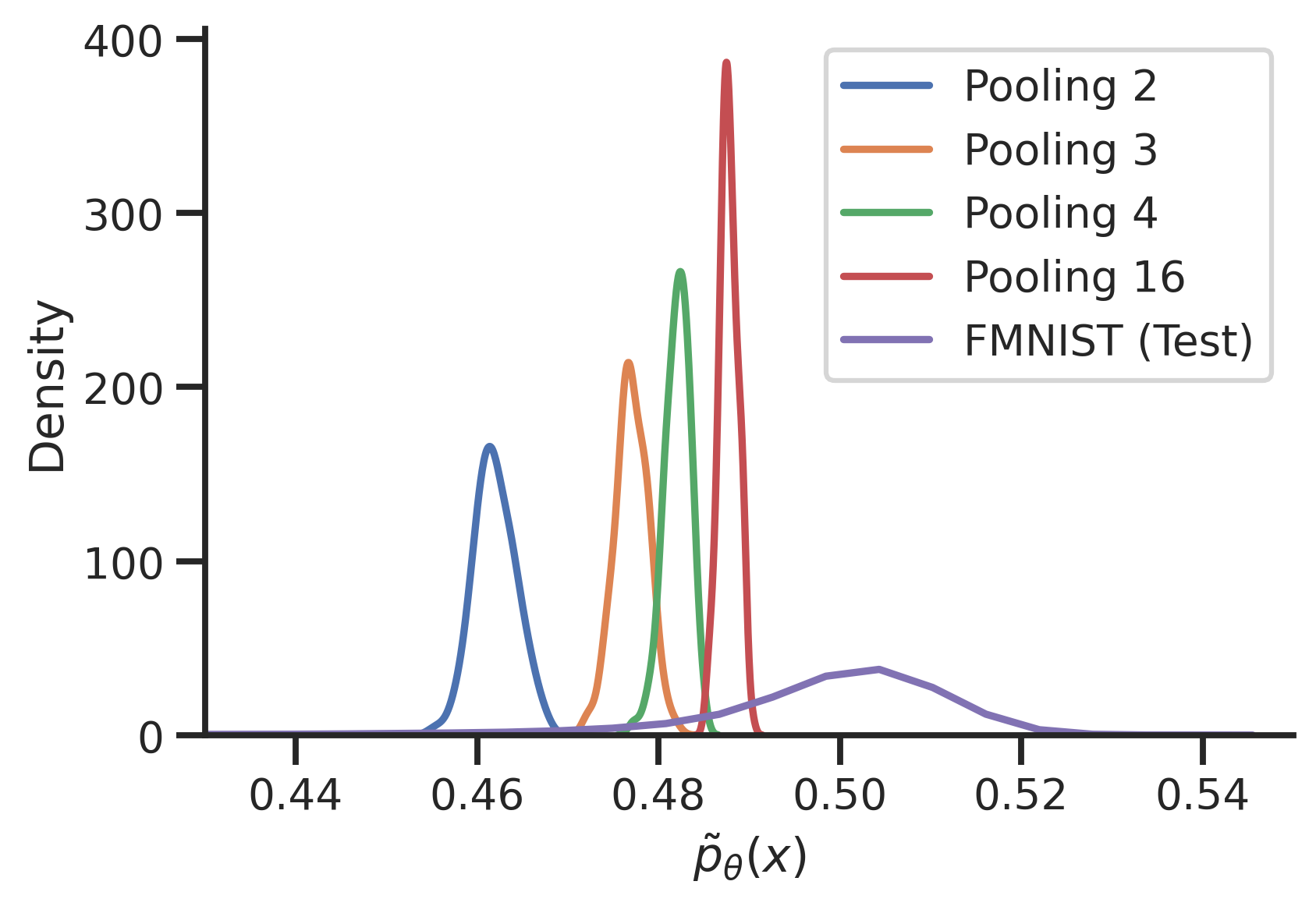}
        \caption{EBM (unsupervised) without bottleneck.}
        \label{fig:density_histogram_no_bottleneck}
    \end{subfigure}%
    \begin{subfigure}[t]{0.45\linewidth}
        \centering
        \includegraphics[width=0.8\linewidth]{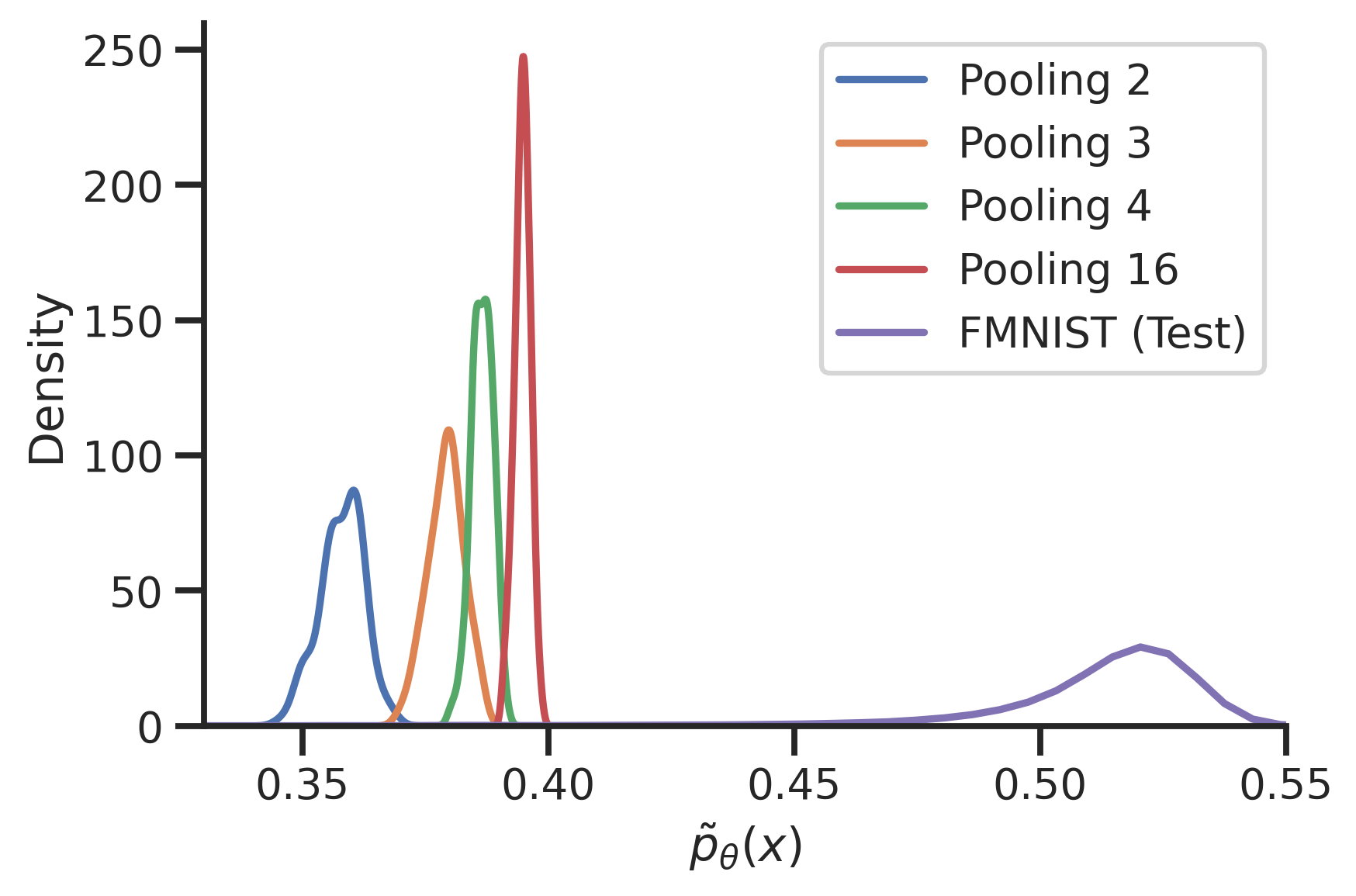}
        \caption{EBM (unsupervised) with bottleneck.}
        \label{fig:density_histogram_with_bottleneck}
    \end{subfigure}%
    \caption{(a, b) Density histograms of generated dataset of noise images with different smoothness under different EBMs. Higher pooling corresponds to higher smoothness of the input images as visualized in \Cref{fig:example_input_images_smooth}. (c, d) Comparison of density histogram of ID test set of FMNIST vs. low-level feature datasets for an EBM with and without bottleneck.}
\end{figure*}

\paragraph{Experiment 5}
To provide further evidence for our observation that low-level features affect the likelihood of unsupervised EBMs, we include density histograms for datasets with varying low-level features. We take inspiration from \cite{serraInputComplexityOutofdistribution2020} and generate images with varying smoothness properties which has shown to affect the likelihood of samples in other generative models. In order to obtain images with different smoothness, we sample uniform noise at each pixel independently, apply average pooling with different pooling sizes, and resize to the original image dimensions with nearest neighbour upsampling. Images after this pre-processing procedure are shown in \Cref{fig:example_input_images_smooth}. Subsequently, we estimate the density of 1000 images generated at each pooling fidelity under our models. Note that we use average pooling in comparison to max pooling in \citet{serraInputComplexityOutofdistribution2020} since max pooling leads to images with different statistics (higher mean) for higher pooling sizes. Average pooling allows use to isolate the contribution of the change of features independent of image statistics. 

In \Cref{fig:density_histogram_smoothness_unsupervised}, we observe that unsupervised EBMs assign higher likelihood to smoother versions of the dataset (corresponding to higher pooling sizes), while the supervised EBM is not affected by the change of low-level features in \Cref{fig:density_histogram_smoothness_supervised}. 
This demonstrates that unsupervised EBMs are susceptible to low-level features affecting the likelihood of samples, while supervised EBMs rely on higher-level, semantic features to assign likelihoods.

In \Cref{fig:density_histogram_no_bottleneck} and \Cref{fig:density_histogram_with_bottleneck}, we investigate the effect of applying the bottleneck to the architecture of the unsupervised EBM. We observe that the EBM with bottleneck assigns higher relative likelihood to the FashionMNIST test set vs. the artificial noise datasets containing low-level features only. This supports our observation in the main paper that including bottlenecks within the EBM helps the model to learn semantic features rather than local, low-level feature correlations.

\begin{figure*}
    \centering
    \begin{subfigure}[t]{0.4\linewidth}
        \centering
        \includegraphics[width=\linewidth]{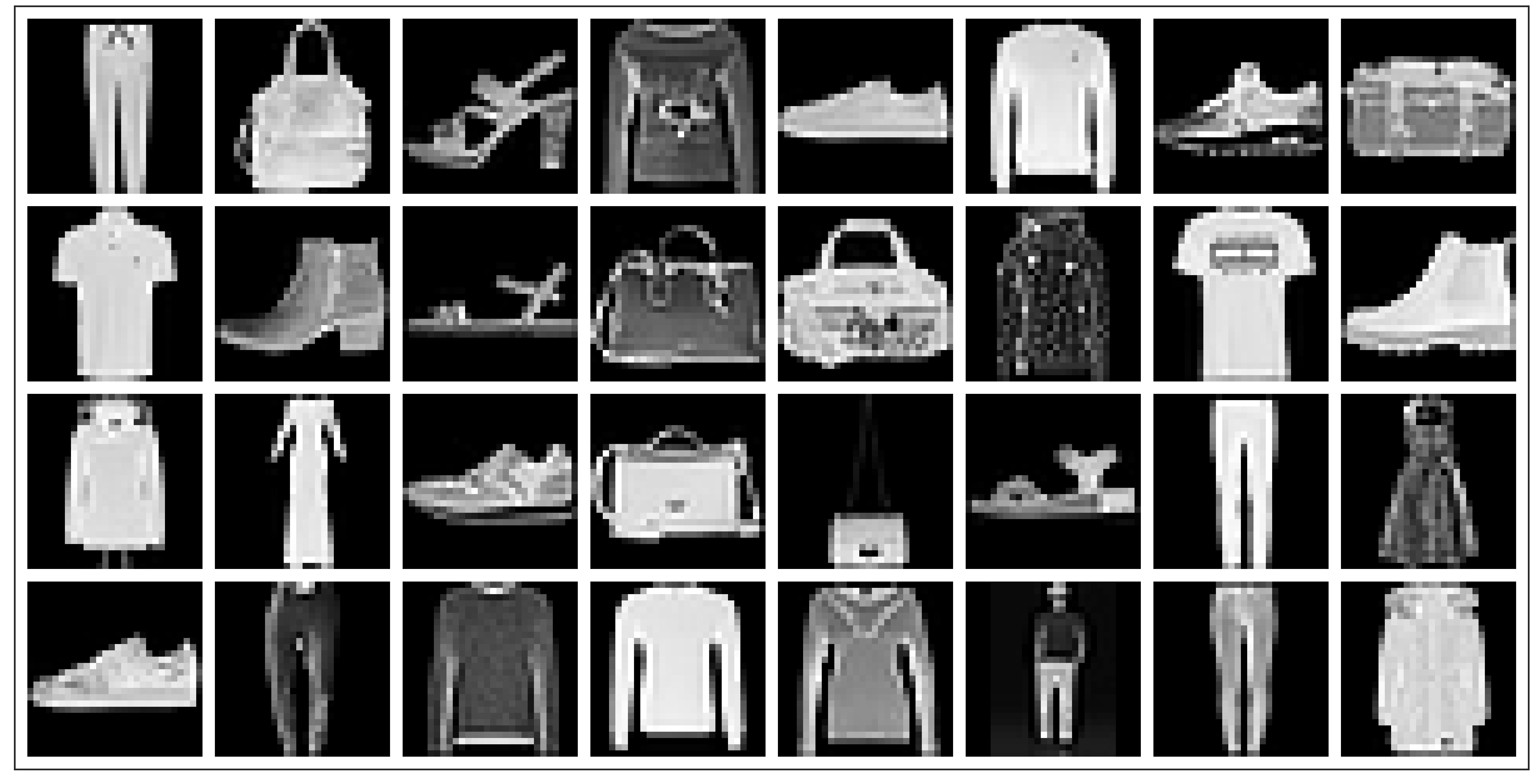}
        \caption{Dataset samples.}
        \label{fig:fmnist_samples}
    \end{subfigure}%
    \hfill
    \begin{subfigure}[t]{0.4\linewidth}
        \centering
        \includegraphics[width=\linewidth]{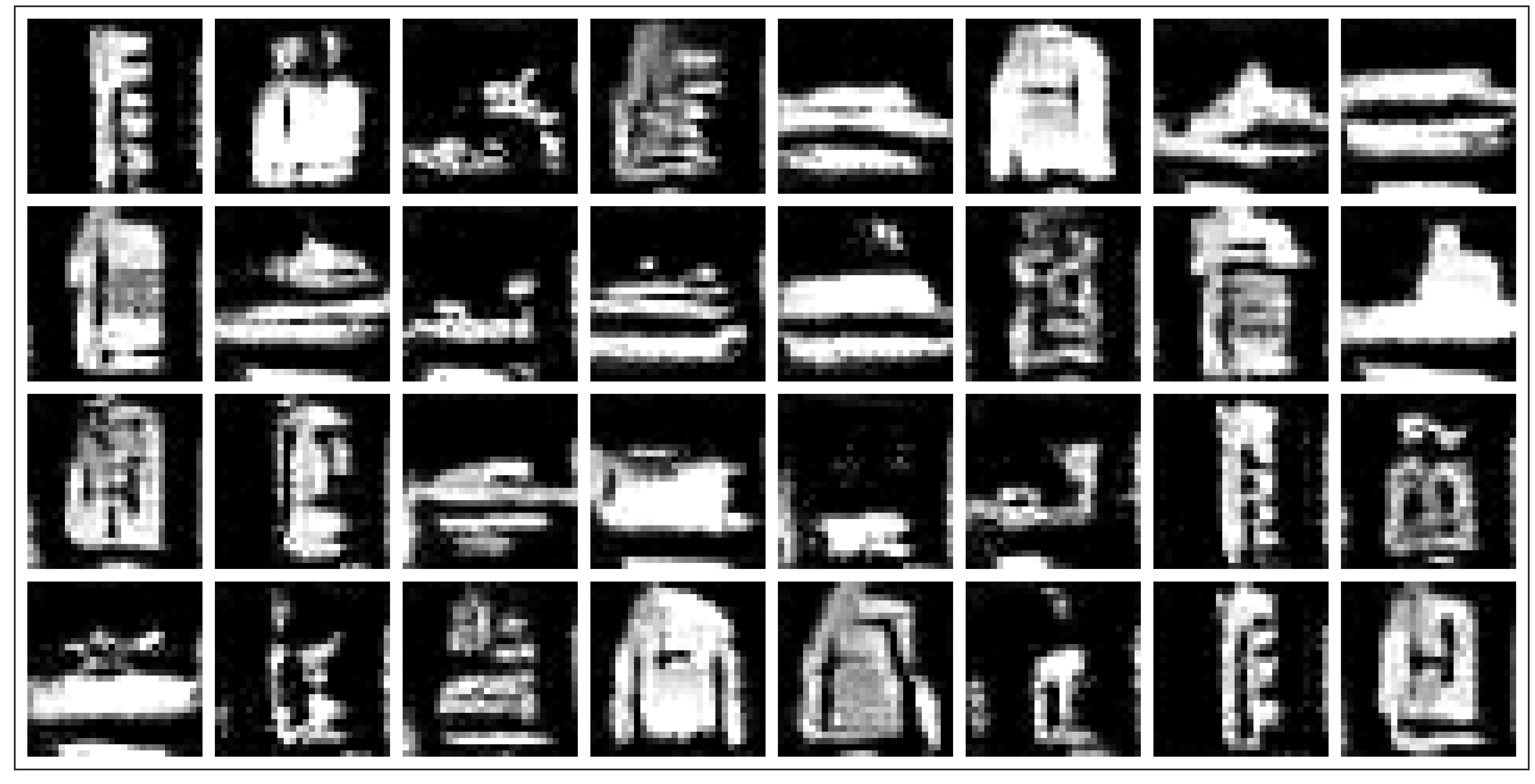}
        \caption{Optimized samples under \textbf{unsupervised} EBM.}
        \label{fig:fmnist_samples_unsupervised}
    \end{subfigure}%
    \\
    \begin{subfigure}[t]{0.4\linewidth}
        \centering
        \includegraphics[width=\linewidth]{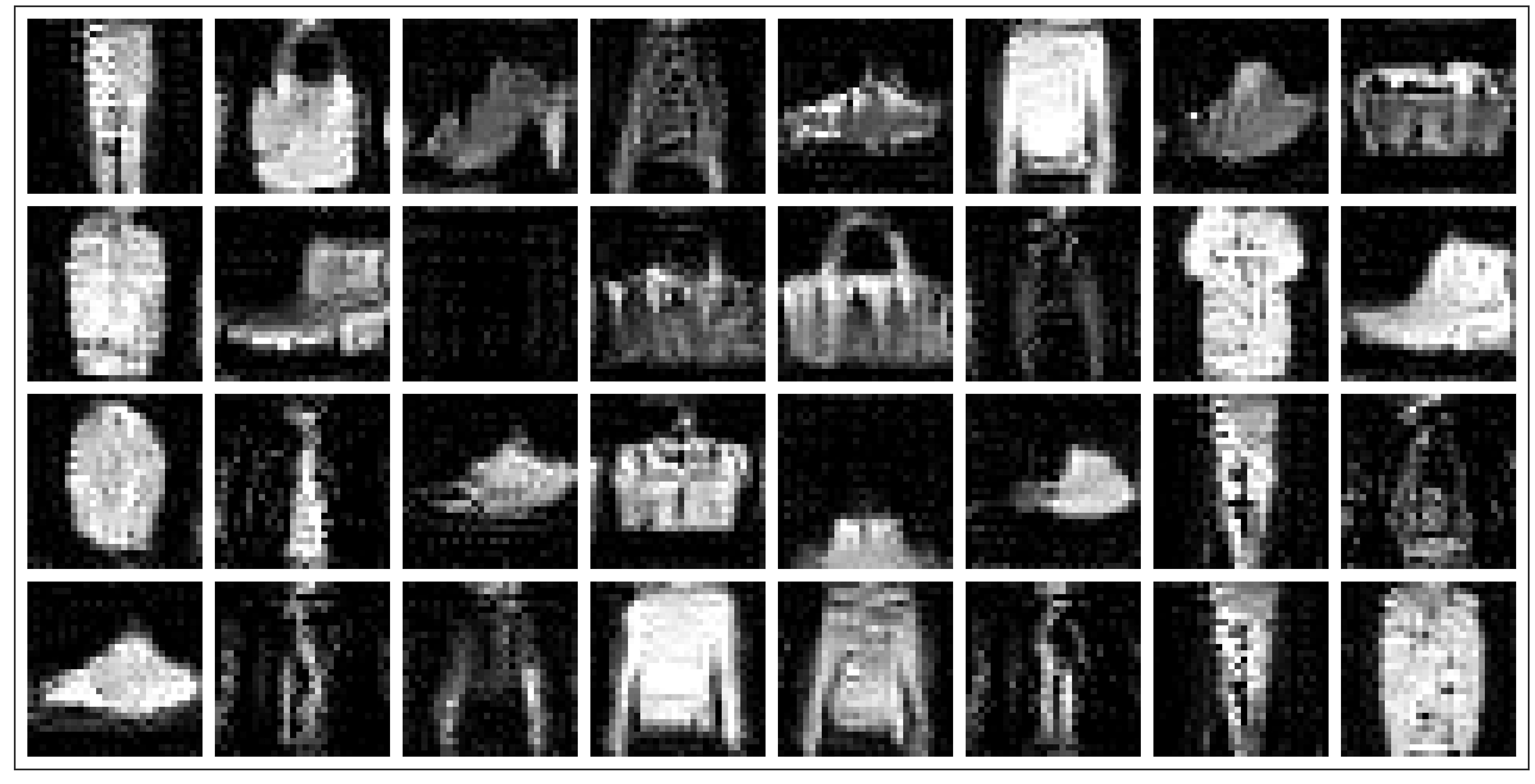}
        \caption{Optimized samples under \textbf{supervised} EBM.}
        \label{fig:fmnist_samples_supervised}
    \end{subfigure}
    \caption{Samples from the FMNIST dataset.}
\end{figure*}

\paragraph{Experiment 6} Finally, we investigate images under the learned EBMs. Samples from the FashionMNIST dataset can be found in \Cref{fig:fmnist_samples}. We optimize the likelihood of these samples under the model and visualize samples for unsupervised EBM in \Cref{fig:fmnist_samples_unsupervised} and for supervised EBM in \Cref{fig:fmnist_samples_supervised}. 

We observe that while the semantic content of samples under the unsupervised model becomes almost indistinguishable, the samples under the supervised model largely preserve their class semantics. 

This result once more highlights that low-level features are the driving factor for high likelihood in unsupervised EBMs, while a notion of semantics is learned in supervised EBMs.

\section{Training details}
\label{sec:training_details}
In this section, we provide further details on the training procedures and hyperparameters used for individual methods. Unless otherwise specified we use the Adam optimizer with default parameters \(\beta_1=0.9\) and \(\beta_2=0.999\). Further, we use learning rate warm-up with \(2500\) steps across all models. We train the models on the tabular datasets for \(10,000\) steps and the image datasets for \(50\) epochs. We perform model selection based on the AUC-PR on an OOD validation dataset. For CIFAR-10, we use the validation sets of CelebA and CIFAR-100, while for FMNIST we use the validation sets of MNIST and KMNIST. On the tabular dataset, we use \(10\%\) of the data of the removed classes for model selection and the remaining data as the test set.

\paragraph{Contrastive Divergence}
Following \cite{grathwohlYourClassifierSecretly2020, duImplicitGenerationGeneralization2020b}, we use persistent contrastive divergence \cite{tielemanTrainingRestrictedBoltzmann2008} which significantly reduces compute compared to seeding new chains at every iteration as in \cite{nijkampLearningNonConvergentNonPersistent2019}. For the parameters of the Stochastic Gradient Langevin Dynamics sampler, we use the settings of \citet{grathwohlYourClassifierSecretly2020} and set the step size \(\alpha\) to \(1\) and reinitialize samples from the replay buffer with probability \(0.05\). The size of the buffer is set to \(10000\). In contrast to \cite{grathwohlYourClassifierSecretly2020}, we found that training with \(20\) SGLD steps consistently diverged, thus, we set the number of SGLD steps to \(100\) which lead to stable convergence. Further, we set the initial learning rate to \(0.001\). We add additive Gaussian noise with variance \(0.1\) to the inputs in order to stabilize training \cite{duImplicitGenerationGeneralization2020b, nijkampLearningNonConvergentNonPersistent2019}.

\paragraph{VERA}
We use the default hyperparameters proposed in \cite{grathwohlNoMCMCMe2020} and initialize the variance of the variational approximation \(\eta\) with \(0.1\) and clamp it in the range \([0.01, 0.3]\). We perform a grid search for the entropy regularizer in \([1e-4, 1]\) and found \(1e^{-4}\) to yield to the best results in terms for training stability.  
Further, we set the learning rate of the EBM to \(3e^{-4}\) and the learning rate of the generator to \(6e^{-4}\). The Adam optimizer is used to optimize the generator with parameters \(\beta_1=0.0\) and \(\beta_2=0.9\).

For the generator architecture, we use a 5-layer MLP for the tabular datasets with hidden dimension \(100\) and leaky ReLU activations with slope \(0.2\). The latent distribution is a \(16\)-dim. isotropic Normal distribution. For the image datasets, we follow \cite{grathwohlNoMCMCMe2020} and use the generator from \cite{miyatoSpectralNormalizationGenerative2018} based on ResNet blocks with latent dimension \(128\).

\paragraph{Sliced Score Matching}
We set the distribution $p_v$ to a multivariate Rademacher distribution which enables to use the variance reduced objective (SSM-VR) where the expectation \(\mathbb{E}_{p_v} \left[ v^T s_m(x;\theta))^2 \right] = \Vert s_m(x;\theta) \Vert_2^2\) can be integrated analytically \cite{songSlicedScoreMatching2019}. \(s_m\) denotes the score function of the EBM. During training, we use a single projection vector \(v\) from $p_v$ to compute the objective.

\paragraph{Normalizing Flow}
We train Normalizing Flow models with maximum likelihood and learning rate \(1e^{-3}\). We perform early stopping based on the log-likelihood with patience \(10\).

\paragraph{Cross-entropy classifier}
We train our cross-entropy baseline with learning rate \(1e^{-3}\) on the tabular and \(1e^{-4}\) on the image datasets. Further, we use weight decay with weight \(5e^{-4}\) and perform early stopping based on the accuracy of the model. 

\section{Architecture details}
\label{sec:architecture_details}
For the Normalizing Flows on the image datasets, we use a Glow \cite{kingmaGlowGenerativeFlow2018} implementation with \(L=3\) layers, \(K=32\) steps, and \(C=512\) channels\footnote{https://github.com/chrischute/glow}. On the tabular datasets, we use \(20\) stacked radial transforms \cite{rezendeVariationalInferenceNormalizing2016}. For all other models, we use a 5-layer MLP with ReLU activations on the tabular datasets and WideResNet-10-2 \cite{zagoruykoWideResidualNetworks2017} on the image datasets.

\section{Dataset details}
In this section, we provide additional details on how we generate \textit{non-natural} OOD datasets used in the paper. For the \textit{Noise} dataset, we use an equal amount of samples from a Gaussian distribution \(N(0, 1)\) and a uniform distribution \(\mathcal{U}(-1, 1)\). The \textit{Constant} dataset is sampled by drawing a scalar from \(\mathcal{U}(-1, 1)\) and then filling a tensor with the same shape as the input data with the sampled value. Finally, \textit{OODomain} inputs are the SVHN dataset and KMNIST dataset, where we do not apply normalization, for the in-distribution datasets of CIFAR-10 and FMNIST, respectively. As a result, the data is in the range \([0, 255]\).

\begin{table*}
    \centering
    \caption{AUC-PR for OOD detection on the natural datasets when trained on the respective in-distribution dataset.}
    \label{tab:full_overall_table}
    \resizebox{\linewidth}{!}{%
    \begin{tabular}{lllllllllll}
\toprule
ID dataset & \multicolumn{5}{c}{CIFAR-10} & \multicolumn{3}{c}{FMNIST} &                                     \multicolumn{1}{c}{Segment} &                                   \multicolumn{1}{c}{Sensorless} \\
\cmidrule(l){2-6} \cmidrule(l){7-9} \cmidrule(l){10-10} \cmidrule(l){11-11} 
OOD dataset &                                   CIFAR-100 &                                       CelebA &                                        LSUN &                                         SVHN &                                    Textures &                                       KMNIST &                                        MNIST &                                     NotMNIST &                                 Segment OOD &                               Sensorless OOD \\
Model  &                                             &                                              &                                             &                                              &                                             &                                              &                                              &                                              &                                             &                                              \\
\midrule
CE     &            62.76 {\footnotesize $\pm$ 1.46} &             64.47 {\footnotesize $\pm$ 2.44} &            65.18 {\footnotesize $\pm$ 5.79} &             47.51 {\footnotesize $\pm$ 4.58} &            39.17 {\footnotesize $\pm$ 2.28} &             69.07 {\footnotesize $\pm$ 6.73} &             82.5 {\footnotesize $\pm$ 12.27} &              50.9 {\footnotesize $\pm$ 6.73} &            33.35 {\footnotesize $\pm$ 1.82} &             33.02 {\footnotesize $\pm$ 1.32} \\
NF     &                                      58.34  &                                       74.68  &                                      62.99  &                                       31.58  &                                      50.23  &                                       62.22  &                                       49.03  &                                       93.68  &                                      99.12  &                                       94.35  \\
\midrule
CD   &            50.51 {\footnotesize $\pm$ 2.13} &             43.86 {\footnotesize $\pm$ 5.85} &           54.43 {\footnotesize $\pm$ 11.37} &            60.72 {\footnotesize $\pm$ 24.59} &           76.21 {\footnotesize $\pm$ 17.44} &             50.52 {\footnotesize $\pm$ 9.39} &              31.69 {\footnotesize $\pm$ 0.9} &             76.85 {\footnotesize $\pm$ 2.66} &            98.18 {\footnotesize $\pm$ 2.18} &            72.83 {\footnotesize $\pm$ 16.19} \\
CD-E &            77.88 {\footnotesize $\pm$ 1.61} &             66.21 {\footnotesize $\pm$ 1.94} &            80.61 {\footnotesize $\pm$ 4.26} &  \bfseries{97.38 {\footnotesize $\pm$ 1.15}} &  \bfseries{98.6 {\footnotesize $\pm$ 0.56}} &             90.53 {\footnotesize $\pm$ 3.38} &             93.05 {\footnotesize $\pm$ 2.88} &              88.05 {\footnotesize $\pm$ 4.5} &                                           - &                                            - \\
CD-S &            49.81 {\footnotesize $\pm$ 1.15} &             50.15 {\footnotesize $\pm$ 4.24} &            53.09 {\footnotesize $\pm$ 1.04} &             45.14 {\footnotesize $\pm$ 7.55} &            46.55 {\footnotesize $\pm$ 4.37} &             93.88 {\footnotesize $\pm$ 1.44} &             83.47 {\footnotesize $\pm$ 2.93} &  \bfseries{98.03 {\footnotesize $\pm$ 0.53}} &  \bfseries{100.0 {\footnotesize $\pm$ 0.0}} &  \bfseries{94.48 {\footnotesize $\pm$ 2.11}} \\
\midrule
SSM    &            53.82 {\footnotesize $\pm$ 3.12} &              57.72 {\footnotesize $\pm$ 7.0} &            52.79 {\footnotesize $\pm$ 3.16} &             45.75 {\footnotesize $\pm$ 7.24} &            48.82 {\footnotesize $\pm$ 4.34} &             58.98 {\footnotesize $\pm$ 5.48} &             67.86 {\footnotesize $\pm$ 11.4} &            57.27 {\footnotesize $\pm$ 13.73} &           79.43 {\footnotesize $\pm$ 24.29} &            67.13 {\footnotesize $\pm$ 20.31} \\
SSM-E  &            84.73 {\footnotesize $\pm$ 0.67} &             78.62 {\footnotesize $\pm$ 2.23} &  \bfseries{89.4 {\footnotesize $\pm$ 1.18}} &             74.69 {\footnotesize $\pm$ 2.76} &             69.8 {\footnotesize $\pm$ 2.72} &             97.88 {\footnotesize $\pm$ 0.66} &             95.57 {\footnotesize $\pm$ 0.97} &             96.44 {\footnotesize $\pm$ 0.83} &                                           - &                                            - \\
SSM-S  &            62.71 {\footnotesize $\pm$ 0.98} &             57.15 {\footnotesize $\pm$ 2.18} &             68.77 {\footnotesize $\pm$ 4.5} &             46.54 {\footnotesize $\pm$ 3.54} &            43.51 {\footnotesize $\pm$ 2.74} &             93.77 {\footnotesize $\pm$ 1.08} &             99.22 {\footnotesize $\pm$ 0.19} &             84.93 {\footnotesize $\pm$ 2.07} &  \bfseries{100.0 {\footnotesize $\pm$ 0.0}} &            81.99 {\footnotesize $\pm$ 21.79} \\
\midrule
VERA   &            55.95 {\footnotesize $\pm$ 2.68} &             73.97 {\footnotesize $\pm$ 2.63} &            67.39 {\footnotesize $\pm$ 2.57} &             37.27 {\footnotesize $\pm$ 4.66} &             46.29 {\footnotesize $\pm$ 8.1} &            78.11 {\footnotesize $\pm$ 21.05} &            67.53 {\footnotesize $\pm$ 21.63} &            76.22 {\footnotesize $\pm$ 22.11} &            94.63 {\footnotesize $\pm$ 7.22} &            45.66 {\footnotesize $\pm$ 10.55} \\
VERA-E &            76.66 {\footnotesize $\pm$ 3.23} &             73.41 {\footnotesize $\pm$ 6.68} &            81.31 {\footnotesize $\pm$ 3.92} &              83.6 {\footnotesize $\pm$ 7.45} &            78.52 {\footnotesize $\pm$ 7.98} &             85.8 {\footnotesize $\pm$ 15.18} &            88.52 {\footnotesize $\pm$ 13.91} &            79.58 {\footnotesize $\pm$ 15.61} &                                           - &                                            - \\
VERA-S &            61.37 {\footnotesize $\pm$ 0.74} &  \bfseries{85.02 {\footnotesize $\pm$ 2.38}} &            58.91 {\footnotesize $\pm$ 3.66} &             38.35 {\footnotesize $\pm$ 1.08} &            36.68 {\footnotesize $\pm$ 0.52} &  \bfseries{98.89 {\footnotesize $\pm$ 0.61}} &  \bfseries{99.64 {\footnotesize $\pm$ 0.53}} &             97.75 {\footnotesize $\pm$ 1.72} &            99.35 {\footnotesize $\pm$ 1.08} &             90.38 {\footnotesize $\pm$ 3.78} \\
\bottomrule
\end{tabular}
    }
\end{table*}

\begin{table*}
    \centering
    \caption{AUC-PR for OOD detection on the non-natural datasets when trained on respective in-distribution dataset.}
    \label{tab:full_overall_table_other}
    \resizebox{\linewidth}{!}{%
    \begin{tabular}{lllllllllll}
\toprule
ID dataset & \multicolumn{3}{c}{CIFAR-10} & \multicolumn{3}{c}{FMNIST} & \multicolumn{2}{c}{Segment} & \multicolumn{2}{c}{Sensorless} \\
\cmidrule(l){2-4} \cmidrule(l){5-7} \cmidrule(l){8-9} \cmidrule(l){10-11} 
OOD dataset &                                     Constant &                                       Noise &                                    OODomain &                                     Constant &                                       Noise &                                    OODomain &                                    Constant &                                       Noise &                                    Constant &                                        Noise \\
Model  &                                              &                                             &                                             &                                              &                                             &                                             &                                             &                                             &                                             &                                              \\
\midrule
CE     &              45.26 {\footnotesize $\pm$ 8.8} &           61.13 {\footnotesize $\pm$ 21.02} &             30.69 {\footnotesize $\pm$ 0.0} &              35.5 {\footnotesize $\pm$ 3.08} &           55.84 {\footnotesize $\pm$ 22.32} &            30.74 {\footnotesize $\pm$ 0.11} &            42.57 {\footnotesize $\pm$ 18.3} &            33.82 {\footnotesize $\pm$ 3.14} &            32.42 {\footnotesize $\pm$ 1.04} &             31.96 {\footnotesize $\pm$ 1.28} \\
NF     &                                       30.87  &            83.65  &                                          \bfseries{100.0} &                                       71.07  &            98.04  &                                          \bfseries{100.0} &                                      99.97  &  \bfseries{100.0} &                           \bfseries{100.0 } &   \bfseries{100.0} \\
\midrule
CD   &            58.75 {\footnotesize $\pm$ 28.17} &  \bfseries{100.0 {\footnotesize $\pm$ 0.0}} &           58.41 {\footnotesize $\pm$ 37.96} &            70.59 {\footnotesize $\pm$ 12.84} &  \bfseries{100.0 {\footnotesize $\pm$ 0.0}} &  \bfseries{100.0 {\footnotesize $\pm$ 0.0}} &            96.13 {\footnotesize $\pm$ 2.55} &            95.43 {\footnotesize $\pm$ 3.58} &  \bfseries{100.0 {\footnotesize $\pm$ 0.0}} &   \bfseries{100.0 {\footnotesize $\pm$ 0.0}} \\
CD-E &  \bfseries{99.92 {\footnotesize $\pm$ 0.07}} &            87.5 {\footnotesize $\pm$ 24.31} &             30.69 {\footnotesize $\pm$ 0.0} &  \bfseries{96.63 {\footnotesize $\pm$ 7.53}} &            86.7 {\footnotesize $\pm$ 26.75} &             35.83 {\footnotesize $\pm$ 7.8} &                                           - &                                           - &                                           - &                                            - \\
CD-S &             31.32 {\footnotesize $\pm$ 0.29} &            98.01 {\footnotesize $\pm$ 0.93} &           70.86 {\footnotesize $\pm$ 30.57} &            71.17 {\footnotesize $\pm$ 10.16} &            97.79 {\footnotesize $\pm$ 1.02} &  \bfseries{100.0 {\footnotesize $\pm$ 0.0}} &            94.57 {\footnotesize $\pm$ 2.11} &             98.68 {\footnotesize $\pm$ 1.9} &            99.97 {\footnotesize $\pm$ 0.06} &             99.98 {\footnotesize $\pm$ 0.03} \\
\midrule
SSM    &            47.24 {\footnotesize $\pm$ 15.56} &           70.28 {\footnotesize $\pm$ 31.39} &            68.57 {\footnotesize $\pm$ 25.2} &            47.57 {\footnotesize $\pm$ 15.18} &           49.45 {\footnotesize $\pm$ 21.19} &            76.76 {\footnotesize $\pm$ 21.1} &           74.86 {\footnotesize $\pm$ 23.37} &           80.35 {\footnotesize $\pm$ 17.38} &            69.66 {\footnotesize $\pm$ 6.12} &            64.81 {\footnotesize $\pm$ 17.03} \\
SSM-E  &             65.64 {\footnotesize $\pm$ 3.66} &             64.5 {\footnotesize $\pm$ 4.05} &            42.74 {\footnotesize $\pm$ 6.73} &             82.51 {\footnotesize $\pm$ 4.95} &            87.54 {\footnotesize $\pm$ 2.17} &            98.49 {\footnotesize $\pm$ 1.94} &                                           - &                                           - &                                           - &                                            - \\
SSM-S  &              37.8 {\footnotesize $\pm$ 3.49} &           64.24 {\footnotesize $\pm$ 12.88} &             30.69 {\footnotesize $\pm$ 0.0} &             37.61 {\footnotesize $\pm$ 1.83} &            33.71 {\footnotesize $\pm$ 1.07} &           70.03 {\footnotesize $\pm$ 17.45} &           51.57 {\footnotesize $\pm$ 25.04} &           70.09 {\footnotesize $\pm$ 17.57} &            37.5 {\footnotesize $\pm$ 13.23} &             41.6 {\footnotesize $\pm$ 12.05} \\
\midrule
VERA   &             31.51 {\footnotesize $\pm$ 0.66} &  \bfseries{100.0 {\footnotesize $\pm$ 0.0}} &           63.48 {\footnotesize $\pm$ 34.37} &            53.24 {\footnotesize $\pm$ 22.65} &           79.34 {\footnotesize $\pm$ 27.34} &           72.42 {\footnotesize $\pm$ 37.61} &  \bfseries{100.0 {\footnotesize $\pm$ 0.0}} &  \bfseries{100.0 {\footnotesize $\pm$ 0.0}} &  \bfseries{100.0 {\footnotesize $\pm$ 0.0}} &             99.88 {\footnotesize $\pm$ 0.25} \\
VERA-E &             83.95 {\footnotesize $\pm$ 8.71} &            36.19 {\footnotesize $\pm$ 4.43} &             30.69 {\footnotesize $\pm$ 0.0} &            77.82 {\footnotesize $\pm$ 21.24} &           60.28 {\footnotesize $\pm$ 10.11} &            60.29 {\footnotesize $\pm$ 28.5} &                                           - &                                           - &                                           - &                                            - \\
VERA-S &              31.7 {\footnotesize $\pm$ 0.55} &           45.32 {\footnotesize $\pm$ 25.14} &  92.1 {\footnotesize $\pm$ 8.59} &             36.01 {\footnotesize $\pm$ 4.24} &            33.73 {\footnotesize $\pm$ 3.21} &  \bfseries{100.0 {\footnotesize $\pm$ 0.0}} &            99.01 {\footnotesize $\pm$ 1.45} &              99.9 {\footnotesize $\pm$ 0.3} &  \bfseries{100.0 {\footnotesize $\pm$ 0.0}} &  \bfseries{100.0 {\footnotesize $\pm$ 0.01}} \\
\bottomrule
\end{tabular}
    }
\end{table*}

\begin{table*}
    \centering
    \caption{AUC-PR for OOD detection of EBMs with different choices of the dimension of the bottleneck introduced into the WideResNet-10-2.
    %\bc{The column order is a bit confusing there} \bc{do we have results for other datasets for this experiment ?}.
    }
    \label{tab:bottleneck_full}
    \resizebox{\linewidth}{!}{%
    \begin{tabular}{llllllllllllllll}
\toprule
     & ID dataset & \multicolumn{8}{c}{CIFAR-10} & \multicolumn{6}{c}{FashionMNIST} \\
\cmidrule(l){3-10} \cmidrule(l){11-16} 
     & OOD dataset &     SVHN &   LSUN & CelebA & CIFAR-100 & Textures &  Noise & OODomain & Constant &       KMNIST &  MNIST & NotMNIST &  Noise & OODomain & Constant \\
Model & Bottleneck &          &        &        &           &          &        &          &          &              &        &          &        &          &          \\
\midrule
\multirow{4}{*}{MCMC} & 0.05 &    81.29 &  42.16 &  42.82 &     53.09 &    59.13 &  100.0 &    76.77 &     78.4 &        65.27 &  43.26 &    80.72 &  100.0 &    100.0 &    73.18 \\
     & 0.10 &    67.93 &   48.3 &  52.71 &     53.47 &    61.99 &  100.0 &    62.19 &    70.85 &         73.8 &  53.23 &    83.25 &  100.0 &    100.0 &    78.27 \\
     & 0.20 &    77.97 &   43.8 &  42.19 &     51.85 &    63.53 &  100.0 &    85.99 &    79.71 &        73.46 &   51.8 &     81.2 &  99.99 &    100.0 &    64.49 \\
     & 1.00 &    60.72 &  54.43 &  43.86 &     50.51 &    76.21 &  100.0 &    58.41 &    58.75 &        50.52 &  31.69 &    76.85 &  100.0 &    100.0 &    70.59 \\
\cline{1-16}
\multirow{4}{*}{SSM} & 0.05 &     53.6 &  49.44 &  54.06 &     53.47 &     43.3 &  51.57 &     89.8 &     67.7 &        56.79 &  62.09 &    49.05 &  42.48 &     77.2 &    46.42 \\
     & 0.10 &    52.51 &   49.4 &   57.4 &     51.54 &    40.74 &  40.81 &    87.05 &    62.91 &        58.32 &  69.04 &    48.75 &  46.52 &    65.72 &    42.82 \\
     & 0.20 &    52.69 &  49.29 &  59.96 &     52.37 &    49.31 &  49.26 &    71.39 &    57.82 &        62.31 &  53.76 &     68.8 &  67.07 &    65.84 &    56.88 \\
     & 1.00 &    45.75 &  52.79 &  57.72 &     53.82 &    48.82 &  70.28 &    68.57 &    47.24 &        58.98 &  67.86 &    57.27 &  49.45 &    76.76 &    47.57 \\
\cline{1-16}
\multirow{4}{*}{VERA} & 0.05 &    33.34 &  89.65 &  82.69 &     55.22 &    58.13 &  72.28 &    86.14 &    30.92 &        87.52 &  80.64 &    75.22 &  74.43 &    100.0 &    33.09 \\
     & 0.10 &    34.52 &  80.64 &   79.4 &     54.44 &    54.84 &  45.02 &    84.56 &    33.65 &        86.85 &  85.65 &     66.3 &  75.58 &    96.28 &    31.54 \\
     & 0.20 &    33.95 &  73.29 &   76.1 &     54.03 &    45.01 &  48.29 &    73.57 &    31.31 &         88.7 &  89.06 &     60.7 &  59.56 &    100.0 &    36.01 \\
     & 1.00 &    37.27 &  67.39 &  73.97 &     55.95 &    46.29 &  100.0 &    63.48 &    31.51 &        78.11 &  67.53 &    76.22 &  79.34 &    72.42 &    53.24 \\
\bottomrule
\end{tabular}
    }
\end{table*}

\end{document}